\documentclass[accepted]{uai2024} 
                        
\usepackage[american]{babel}

\usepackage{natbib} 
\usepackage{mathtools} 
\usepackage{booktabs} 
\usepackage{tikz} 

\font\tenrm = cmr17 at 10pt 

\usepackage{macros}

\usepackage{color}

\title{Efficient Sensor Placement from Regression with Sparse Gaussian Processes}

\author[1]{\href{mailto:kjakkala@charlotte.edu}{Kalvik~Jakkala}{}}
\author[1]{Srinivas~Akella}
\affil[1]{%
    Computer Science Dept.\\
    University of North Carolina at Charlotte\\
    Charlotte, North Carolina, USA
}

\begin{document}
\maketitle

\begin{abstract}
The sensor placement problem is a common problem that arises when monitoring correlated phenomena, such as temperature, precipitation, and salinity. Existing approaches to this problem typically formulate it as the maximization of information metrics, such as mutual information~(MI), and use optimization methods such as greedy algorithms in discrete domains, and derivative-free optimization methods such as genetic algorithms in continuous domains. However, computing MI for sensor placement requires discretizing the environment, and its computation cost depends on the size of the discretized environment. These limitations restrict these approaches from scaling to large problems.

We present a novel formulation to the SP problem based on variational approximation that can be optimized using gradient descent, allowing us to efficiently find solutions in continuous domains. We generalize our method to also handle discrete environments. Our experimental results on four real-world datasets demonstrate that our approach generates sensor placements consistently on par with or better than the prior state-of-the-art approaches in terms of both MI and reconstruction quality, all while being significantly faster. Our computationally efficient approach enables both large-scale sensor placement and fast robotic sensor placement for informative path planning algorithms.
\end{abstract}
\section{Introduction}

Meteorology and climate change are concerned with monitoring correlated environmental phenomena such as temperature, ozone concentration, soil chemistry, ocean salinity, and fugitive gas density~\citep{krauseSG08, MaLS17, SuryanT20, WhitmanMKC21, JakkalaA22}. However, it is often too expensive and, in some cases, even infeasible to monitor the entire environment with a dense sensor network. We therefore aim to determine strategic locations for a sparse set of sensors so that the data from these sensors gives us the most accurate estimate of the phenomenon over the entire environment. We address this \textit{sensor placement problem} for correlated environment monitoring.

Moreover, we focus on the sparsely labeled sensor placement problem, wherein only a few labeled data samples are available. This data restriction limits the applicability of many parametric approaches~\citep{GoodfellowBC16}, such as deep learning and deep reinforcement learning. The sparsely labeled sensor placement problem is a fundamental problem with diverse and important applications. For example, informative path planning (IPP) is a crucial problem in robotics that involves identifying informative sensing locations for robots while considering travel distance constraints~\citep{MaLS17}. Similar sensor placement problems arise in autonomous robot inspection and monitoring of 3D surfaces~\citep{ZhuCLSA21}, for example, when a robot must monitor stress fractures on an aircraft body. Recently a sensor placement approach has even been used to learn dynamical systems in a sample-efficient manner~\citep{FenetST20}. These problems require fast solutions, but the sensor placement problem often becomes a computationally expensive bottleneck in current approaches.

An effective approach to address the considered sensor placement problem is to use Gaussian processes (GPs)~\citep{HusainC80, ShewryW87, WuZ92, krauseSG08}. We can capture the correlations of the environment using the GP's kernel function and then leverage the GP to estimate information metrics such as mutual information (MI). Such metrics can be used to quantify the amount of new information that can be obtained from each candidate sensor location. However, computing MI using GPs is very expensive as it requires the inversion of large covariance matrices whose size increases with the environment's discretization resolution. Consequently, these methods do not scale well to continuous spaces and 3D spaces, and have limited applicability when addressing the aforementioned applications, which often necessitate a large number of sensor placements or a fine sensor placement precision that is infeasible with discrete approaches.

Sparse Gaussian processes (SGPs) \citep{CandelaRW07} are a computationally efficient variant of GPs. Therefore, one might consider using SGPs instead of GPs in GP-based sensor placement approaches. However, a naive replacement of GPs with SGPs is not always possible or efficient. This is because SGPs must be retrained for each evaluation of MI. In sensor placement approaches, MI is often evaluated repeatedly, making SGPs computationally more expensive than GPs for sensor placement. So even though SGPs have been studied for over two decades, SGPs have received limited attention for addressing the sensor placement problem~\citep{BigoniZH20}.

The objective of this paper is to develop an efficient approach for addressing the sparsely labeled sensor placement problem in correlated environments. We present an efficient, sparsely supervised, gradient-based approach for sensor placement in continuous environments by formulating the problem using variational inference. Unlike most prior approaches, our method is fully differentiable. As such, it can even be incorporated into deep neural networks optimized for other downstream tasks. We also generalize our method to efficiently handle sensor placement in discrete environments. Our approach enables efficient sensor placement in 3D spaces, spatiotemporally correlated spaces, and derivative problems such as informative path planning critical for robotics applications.
\section{Problem Statement}
\label{sec:problem_statement}
Consider a correlated stochastic process $\Psi$ over an environment $\mathcal{V} \subseteq \mathbb{R}^d$ modeling a phenomenon such as temperature. The \textit{sensor placement problem} is to select a set $\mathcal{A}$ of $s$ sensor locations $\{\mathbf{x}_i \in \mathcal{V}, i = 1,...,s\}$ so that the data $y_i \in \mathbb{R}$ collected at these locations gives us the most accurate estimate of the phenomenon at every location in the environment. We consider estimates with the lowest root-mean-square error (RMSE) to be the most accurate. An ideal solution to this sensor placement problem should have the following key properties:
\begin{enumerate}
    \item The approach should be computationally efficient and produce solutions with low RMSE. Since the environment is correlated, this should also result in the solution sensor placements being well separated to ensure that the sensors collect only novel data that is crucial for accurately reconstructing the data field.    
    \item The approach should handle both densely and sparsely labeled environments. In a densely labeled environment, we have labeled data at every location in the environment.  In a sparsely labeled environment, we have labeled data that is sufficient only to capture the correlations in the environment, or we have domain knowledge about how the environment is correlated.
    \item The approach should handle both continuous sensor placements $\mathcal{A} \subseteq \mathcal{V}$, where the sensors can be placed anywhere in the environment, and discrete sensor placements $\mathcal{A} \subseteq \mathcal{S} \subseteq \mathcal{V}$, where the sensors can only be placed at a subset of a pre-defined set of locations $\mathcal{S}$.
\end{enumerate}
\section{Related Work}
Early approaches to the sensor placement problem~\citep{BaiKXYL06, Ramsden09} used geometric models of the sensor's field of view to account for the region covered by each sensor and used computational geometry or integer programming methods to find solutions. Such approaches proved useful for problems such as the art gallery problem~\citep{DeBergCKO08}, which requires one to place cameras so that the entire environment is visible. However, these approaches do not consider the spatial correlations in the environment.

This problem is also studied in robotics \citep{CortesMKB04, BreitenmoserSMSR10, SadeghiAS22}. Similar to geometric approaches, authors focus on coverage by leveraging Voronoi decompositions~\citep{DeBergCKO08}. A few authors \citep{SchwagerVPRT17, SalamH19}, have even considered Gaussian kernel functions, but they did not leverage the full potential of Gaussian processes.

Gaussian process (GP) based approaches addressed the limitations of geometric model-based sensor placement approaches by learning the spatial correlations in the environment. The learned GP is then used to quantify the information gained from each sensor placement while accounting for the correlations of the data field. However, these methods require one to discretize the environment and introduce severe computational scaling issues. Our method finds sensor locations in continuous spaces and overcomes the computational scaling issues.

Early GP-based approaches~\citep{ShewryW87, WuZ92} placed sensors at the highest entropy locations. However, since GPs have high variance in regions of the environment far from the locations of the training samples, such approaches tended to place sensors at the sensing area's borders, resulting in poor coverage of the area of interest. \citep{krauseSG08} used mutual information (MI) computed with GPs to select sensor locations with the maximal information about all the unsensed locations in the environment. The approach avoided placing the sensors at the environment's boundaries and outperformed all earlier approaches in terms of reconstruction quality and computational cost.

\citet{TajnafoiAMVMPG21} leveraged a full variational Gaussian process to compute mutual information (MI) and then utilized a stochastic greedy algorithm to select the solution sensors that maximize MI. This method is a computationally faster generalization of \citet{krauseSG08}. However, it still discretizes the environment and performs a combinatorial search, which limits its scalability to problems in 3D spaces.

\citet{WhitmanMKC21} recently proposed an approach to model spatiotemporal data fields using a combination of sparse Gaussian processes~(SGPs) and state space models. They then used the spatiotemporal model to sequentially place sensors in a discretized version of the environment. Although their spatiotemporal model of the environment resulted in superior sensor placements, the combinatorial search becomes prohibitively large and limits the size of the problems that can be solved using their method.

\citet{LinCWT19} addressed sensor placement in continuous spaces by first mapping the data samples to a lower-dimensional space and then estimating a lower bound for the mutual information among the data. The lower bound on the mutual information was then optimized using Bayesian optimization to find the solution sensing locations. Sensor placement in continuous environments has also been addressed in the context of informative path planning using gradient-free optimization methods such as evolutionary algorithms~\citep{HitzGGPS17} and Bayesian optimization~\citep{FrancisOMR19}. However, both of these approaches maximize MI computed using GPs, which is computationally expensive. In addition, evolutionary algorithms and Bayesian optimization are known to have poor scalability.

A closely related problem is sensor placement with labeled data, where we either have a large corpus of real or simulated data, which is used to select a subset of sensing locations and validate their performance from the reconstruction error on the training set. \citet{Semaan17} used random forests, \citet{BigoniZH20} used SGPs and optimized them using genetic algorithms, \citet{WangLC20} leveraged reinforcement learning, and \citet{LiLW21} used linear programming to address this problem. Although such approaches are capable of finding good sensor placement locations, they do not generalize to novel environments where we do not have any significant amounts of data. Our approach addresses the sparsely labeled sensor placement problem.
\section{Gaussian processes}

Gaussian processes (GPs) \citep{RasmussenW05} are a non-parametric Bayesian approach that we can use for regression, classification, and generative problems. Suppose we are given a regression task's training set $\mathcal{D} = \{(\mathbf{x}_i, y_i), i = 1,...,n\}$ with $n$ data samples consisting of inputs $\mathbf{x}_i \in \mathbb{R}^d$ and noisy labels $y_i \in \mathbb{R}$, such that,  $y_i = f(\mathbf{x}_i) + \epsilon_i$, where $\epsilon_i \sim \mathcal{N}(0, \sigma^2_{\text{noise}})$. Here $\sigma^2_{\text{noise}}$ is the variance of the independent additive Gaussian noise in the observed labels $y_i$, and the latent function $f(\mathbf{x})$ models the noise-free function of interest that characterizes the regression dataset.

GPs model such datasets by assuming a GP prior over the space of functions that we could use to model the dataset, i.e., they assume the prior distribution over the function of interest $p(\mathbf{f} | \mathbf{X}) = \mathcal{N}(0, \mathbf{K})$, where $\mathbf{f} = [f_1, f_2,...,f_n]^\top$ is a vector of latent function values, $f_i = f(\mathbf{x}_i)$. $\mathbf{X} = [\mathbf{x}_1, \mathbf{x}_2,...,\mathbf{x}_n]^\top$ is a vector (or matrix) of inputs, and $\mathbf{K} \in \mathbb{R}^{n \times n}$ is a covariance matrix, whose entries $\mathbf{K}_{ij}$ are given by the kernel function $k(\mathbf{x}_i, \mathbf{x}_j)$. 

The kernel function parameters are tuned using Type II maximum likelihood~\citep{Bishop06} so that the GP accurately predicts the training dataset labels. \comment{We can compute the posterior of the GP with the mean and covariance functions:}
\comment{\begin{equation}
\begin{aligned}
\label{eq:gp}
m_\mathbf{y}(\mathbf{x}) &= \mathbf{K}_{xn}(\mathbf{K}_{nn} + \sigma_{\text{noise}}^{2}I)^{-1}\mathbf{y}\,, \\
k_\mathbf{y}(\mathbf{x}, \mathbf{x}^\prime) &= k(\mathbf{x}, \mathbf{x}^\prime) - \mathbf{K}_{xn}(\mathbf{K}_{nn} + \sigma_{\text{noise}}^{2}I)^{-1}\mathbf{K}_{nx^\prime}\,,
\end{aligned}
\end{equation}}
\comment{where $\mathbf{y}$ is a vector of all the outputs, and the covariance matrix subscripts indicate the variables used to compute it, i.e., $\mathbf{K}_{nn}$ is the covariance of the training inputs $\mathbf{X}$, and $\mathbf{K}_{xn}$ is the covariance between the test input $\mathbf{x}$ and the training inputs $\mathbf{X}$.} This approach requires an inversion of a matrix of size $n \times n$, which is a $\mathcal{O}(n^3)$ operation, where $n$ is the number of training set samples. Thus this method can handle at most a few thousand training samples. 
\section{Method}

\subsection{Theoretical Foundation}
We address the sensor placement~(SP) problem by treating the underlying data field in the environment as a stochastic process that can be modeled using a Gaussian process~(GP). Note that GPs are not limited to modeling Gaussian distributed functions; a GP is a set of random variables $\mathbf{f(X)} = \{f(\mathbf{x}) \mid \mathbf{x} \in \mathbf{X} \}$ for which any finite subset follows a Gaussian distribution. Here, the latent variables $\mathbf{f(X)}$ represent the noise-free labels of the phenomenon being monitored at every location in the environment.

To find the solution sensor placements, we approximate the abovementioned GP using a sparse distribution $q$, which introduces $m$ auxiliary inducing latent variables $\mathbf{f}_m(\mathbf{X}_m)$:

\begin{equation}
\begin{aligned}
    \hat{\mathbf{X}} &= \{ \mathbf{X}, \mathbf{X}_m \} \,, \\
    q(\hat{\mathbf{f}}(\hat{\mathbf{X}})) &= p(\mathbf{f(X)}, \mathbf{f}_m(\mathbf{X}_m)) \\
    &= p(\mathbf{f(X)} | \mathbf{f}_m(\mathbf{X}_m)) q(\mathbf{f}_m(\mathbf{X}_m)) \,.
\end{aligned}
\end{equation}

Here, the sparse distribution $q$ is formulated so that all the information relevant to predicting the latent variables $\mathbf{f}$ is captured by the distribution $ q(\mathbf{f}_m(\mathbf{X}_m))$, i.e., knowing the latents $\mathbf{f}_m(\mathbf{X}_m)$ would suffice to predict all remaining latent variables $\mathbf{f(X)}$. Both $\mathbf{X}$ and $\mathbf{X}_m$ are drawn from a Gaussian process prior; as such, the conditional $p(\mathbf{f(X)} | \mathbf{f}_m(\mathbf{X}_m))$ can be explicitly computed for a given $\mathbf{X}$ and $\mathbf{X}_m$. Therefore, only $q(\mathbf{f}_m(\mathbf{X}_m))$ needs to be optimized in $q(\hat{\mathbf{f}}(\hat{\mathbf{X}}))$. Note that from here on, we stop explicitly denoting the dependence of the latent variables $\mathbf{f}$ on the input locations $\mathbf{X}$ to simplify notation.

We optimize the sparse distribution $q(\hat{\mathbf{f}})$ by minimizing the KL-divergence between the full GP modeling the underlying process $p(\hat{\mathbf{f}})$ and the sparse distribution $q(\hat{\mathbf{f}})$:

\begin{equation}
\label{eq:forward_kl}
    \mathcal{F}(q) = \text{KL}(p(\hat{\mathbf{f}})||q(\hat{\mathbf{f}}))
\end{equation}

But $p(\mathbf{\hat{f}})$ in the above formulation is only a prior on the stochastic process modeling the phenomenon of interest, and we need to provide it with labeled data $\mathbf{y}$, i.e., $p(\mathbf{\hat{f}|y}))$ for it to represent the current state of the process. Including $\mathbf{y}$ has two key implications. First, it allows us to learn the GP hyperparameters (i.e., the kernel function parameters and data noise variance), thereby capturing the correlations in the stochastic process. Second, it biases our optimized sparse approximation to ensure that it captures the current realization of the stochastic process at the locations corresponding to the labels $\mathbf{y}$.

However, since our SP problem does not assume access to the ground truth labeled data, we cannot include it in our formulation. We address this problem by introducing an uninformative training set with the labels $\mathbf{y}$ all set to zero, and the inputs are defined to be the locations within the boundaries of the environment. Such an approach would still bias the sparse approximation to match the process within the confines of the monitoring environment. But we cannot use it to learn the hyperparameters. We address this issue by learning the hyperparameters from the sparse set of labeled data provided to us, as mentioned in the problem statement. We optimize the hyperparameters using the standard GP formulation and type II maximum likelihood~\citep{RasmussenW05}.

We now address optimizing the sparse approximation~$q$. The current formulation~(Equation~{\ref{eq:forward_kl}}), which uses the forward KL-divergence, is computationally intractable to optimize if we include the labels~\citep{BleiKM17}. Nonetheless, there are multiple alternative approaches to optimize the sparse approximation, such as variational inference and expectation propagation~\citep{Bishop06}. We leverage variational inference here, which gives us the following evidence lower bound~(ELBO) as the new optimization objective:

\begin{equation}
    \mathcal{F}(q) = \mathbb{E}_q \left[ \log p({\mathbf{\hat{f}, y}}) \right] - \mathbb{E}_q \left[ \log q({\mathbf{\hat{f}}}) \right] \,.
\end{equation}

Note that the above uses the reverse KL-divergence, and the training set labels $\mathbf{y}$ are all set to zero, with the corresponding inputs being the locations from within the monitoring environment. Substituting the full GP and the sparse approximation into the above ELBO gives us the following:

\begin{equation}
\begin{aligned}
    \mathcal{F}(q) &= \int q(\mathbf{f}_m) \log \frac{\mathcal{N}(\mathbf{y} | \boldsymbol{\alpha}, \sigma_{\text{noise}}^2I) p(\mathbf{f}_m)}{q(\mathbf{f}_m)} d\mathbf{f}_m \\
    & \quad - \frac{1}{2 \sigma_{\text{noise}}^2} Tr (\mathbf{K}_{ff} - \mathbf{Q}) \\
    \boldsymbol{\alpha} &= \mathbb{E}[\mathbf{f} \mid \mathbf{f}_m] = \mathbf{K}_{nm}\mathbf{K}_{mm}^{-1}\mathbf{f}_m \\
\mathbf{Q} &= \mathbf{K}_{nm}\mathbf{K}_{mm}^{-1}\mathbf{K}_{mn} \,.
\end{aligned}
\end{equation}

Here, $\sigma_{\text{noise}}$ is the data noise, and the covariance matrix $\mathbf{K}$ subscripts indicate the variables used to compute it, with $n$ indicating the $n$ zero-labeled points $\mathbf{X}$ corresponding to $\mathbf{y}$ and $m$ indicating the $m$ inducing points $\mathbf{X}_m$. By leveraging Jensen's inequality~\citep{Bishop06}, we get the following optimal sparse distribution $q^*$ as the solution to the ELBO:

\begin{equation}
    q^*(\hat{\mathbf{f}}) = \mathcal{N}(\mathbf{f}_m|\sigma_{\text{noise}}^{-2}\mathbf{K}_{mm}\mathbf{\Sigma}^{-1}\mathbf{K}_{mn}\mathbf{y}, \mathbf{K}_{mm}\mathbf{\Sigma}^{-1}\mathbf{K}_{mm}) \,,
\end{equation}

where $\mathbf{\Sigma} = \mathbf{K}_{mm} + \sigma_{\text{noise}}^{-2}\mathbf{K}_{mn}\mathbf{K}_{nm}$. The input locations $\mathbf{X}_m$ corresponding to the auxiliary latents $\mathbf{f}_m$ used in the optimal sparse approximation can be optimized by maximizing the following expanded form of the ELBO:

\begin{equation}
\label{vfe}
\begin{aligned}
\mathcal{F}(q) &= \underbrace{\frac{n}{2} \log(2\pi)}_{\text{constant}} + \underbrace{\frac{1}{2} \mathbf{y}^\top (\mathbf{Q}_{nn} + \sigma_{\text{noise}}^2 I)^{-1} \mathbf{y}}_{\text{data fit}} + \\
& \underbrace{\frac{1}{2} \log |\mathbf{Q}_{nn} + \sigma_{\text{noise}}^2 I|}_{\text{complexity term}} - \underbrace{\frac{1}{2\sigma_{\text{noise}}^2} Tr(\mathbf{K}_{nn} - \mathbf{Q}_{nn})}_{\text{trace term}}\,, \\
\end{aligned}
\end{equation}

where $\mathbf{Q}_{nn} = \mathbf{K}_{nm} \mathbf{K}_{mm}^{-1} \mathbf{K}_{mn}$. The lower bound $\mathcal{F}$ has three key terms. The data fit term ensures that the training set labels are accurately predicted. Here, the data fit term is disabled by setting the labels to zero. The complexity and trace terms are independent of the labels. The complexity term ensures that the inducing points (i.e., the input points $\mathbf{X}_m$ corresponding to the $m$ auxiliary latent variables $\mathbf{f}_m$) are spread apart to ensure good coverage of the whole training set, and the trace term represents the sum of the variance of the conditional $p(\mathbf{f}|\mathbf{f}_m)$. When the trace term becomes zero, the $m$ solution inducing points become a sufficient statistic for the $n$ training samples, i.e., the $m$ solution inducing points can make the same predictions as a GP with all the $n$ samples in its training set.

The optimized $\mathbf{X}_m$ correspond to the solution sensor placement locations since the data collected at these locations can be used to recover the state of the full GP modeling the phenomenon of interest. Also, the inducing points can be optimized in continuous spaces using gradient-based approaches such as gradient descent and Newton's method. In contrast to the non-differentiable SP approaches, the above formulation does not require one to discretize the environment, and is significantly faster to optimize. As such, the formulation scales well to large sensor placement problems.

Moreover, the above result can be viewed as a special case of sparse Gaussian processes~(SGPs, \cite{SnelsonG06, Titsias09}) with the training set inputs corresponding to the locations within the data field being monitored and the labels all set to zero. In particular, our formulation, therefore, closely follows that of the sparse variational free energy-based Gaussian process~(SVGP); please refer to \cite{Titsias09} for additional details of their derivation.

We started with the sensor placement problem, treating the data field of interest as a stochastic process modeled by a Gaussian process. Next, we presented an approximate inference-based formulation of the problem and derived the solution using variational inference. We then connected this to the SGP literature by noting that SGPs essentially solve the sensor placement problem in a supervised manner. Furthermore, our formulation demonstrates that when the kernel hyperparameters are known, SGPs can be trained in an unsupervised manner by setting the training set labels to zero and optimizing the sensor placement locations.

Our formulation also enables leveraging the vast SGP literature to address multiple variants of the SP problem. For instance, we can use stochastic gradient optimizable SGPs~\citep{HensmanFL13, wilkinsonSS21} with our approach to address significantly large SP problems, i.e., environments that require a large number of sensor placements and have numerous obstacles that need to be avoided. Similarly, we can use spatiotemporal SGPs~\citep{HamelijnckWLSD21} with our approach to efficiently optimize sensor placements for spatiotemporally correlated environments. \cite{BauerWR16} conducted an in-depth analysis of the SVGP's lower bound; we can even leverage these findings to understand how to better optimize the inducing points in our formulation. 

Additionally, this formulation enables us to lower bound the KL divergence between the sparse approximation and the ground truth full GP modeling the underlying phenomenon in the environment. This lower bound can be used to assess the quality of the sparse approximation computed from the solution sensor placements:

\begin{theorem}\normalfont\citep{BurtRW20}
\label{thm:bound}
 Suppose $N$ training inputs are drawn i.i.d according to input density $p(\mathbf{x})$, and $k(\mathbf{x, x}) < v$ for all $\mathbf{x} \in \mathbf{X}$. Sample $M$ inducing points from the training data with the probability assigned to any set of size $M$ equal to the probability assigned to the corresponding subset by an $\epsilon$ k-Determinantal Point Process with k~$ = M$. With probability at least $1 -\delta$,
\
$$KL(Q||\hat{P}) \leq \frac{C(M+1)+2Nv\epsilon}{2\sigma^2_n \delta}\left( 1+\frac{||\mathbf{y}||^2_2}{\sigma^2_n} \right)$$

where $C = N \sum^\infty_{m=M+1} \lambda_m$, $\lambda_m$ are the eigenvalues of the integral operator $\mathcal{K}$ for kernel $k$ and $p(\mathbf{x})$.
\end{theorem}

Here, k represents the k-value in the k-determinantal point process, and $k$ represents the kernel function. In our sensor placement problem, $Q$ is equivalent to the sparse approximation that can be used to predict the state of the whole environment from sensor data collected at the inducing points, and $\hat{P}$ is the ground truth GP that senses every location in the environment. The theorem also suggests an asymptotic convergence guarantee, i.e., as the number of sensing locations increases, the probability of the sparse approximation being exact approaches one.

Please refer to Appendix~\ref{sec:theory} for additional theoretical analysis of the approach, which details our derivation to show that our formulation is not submodular.

\subsection{Continuous-SGP: Continuous Space Solutions}
\label{Continuous-SGP}

We now detail Algorithm~\ref{alg:Continuous-SGP}, the Continuous-SGP algorithm that leverages the formulation above to address the SP problem in continuous spaces. Given an environment, we first sample random unlabeled locations within the boundaries of the monitoring regions. These unlabeled locations $\mathbf{X}$ are used as the training set inputs, and their labels are all set to zero. Note that the number of inducing points here is represented by $s$.

\begin{algorithm}[hb]
\caption{Continuous-SGP approach for obtaining sensor placements in continuous spaces. Here, $\theta$ are the hyperparameters learned from either historical data or expert knowledge, $\Phi$ is a random distribution defined within the boundaries of the environment $\mathcal{V}$, $s$ is the number of required sensors, $n$ is the number of random unlabeled locations used to train the SGP, and $\gamma$ is the SGP learning rate.}
\label{alg:Continuous-SGP}
\SetKwInput{kwLoop}{Loop until}
\KwIn{$\theta, \mathcal{V}, \Phi$, $s$, $n$, $\gamma$}
\KwOut{Sensor placements $\mathcal{A} \in \mathcal{V}$, where $|\mathcal{A}| = s$}

$\mathbf{X} \sim \Phi(\mathcal{V})$ \ \text{/ / \tenrm{Draw $n$ unlabeled locations}}

$\mathbf{X}_m \sim \Phi(\mathcal{V})$ \ \text{/ / \tenrm{Draw $s$ inducing point locations}}

\tcc{Initialize the SGP with zero mean and zero labeled data}
 
$\varphi = \mathcal{SGP}(\text{mean}=0, \theta; \mathbf{X}, \mathbf{y}=\mathbf{0}, \mathbf{X}_m)$

\tcc{Optimize the inducing points $\mathbf{X}_m$ by maximizing the objective function $\mathcal{F}$ of the SGP $\varphi$ using gradient descent with a learning rate of~$\gamma$}
$\textbf{Loop until} \textit{ convergence}: $

$\quad \quad \quad \mathbf{X}_m \leftarrow \mathbf{X}_m + \gamma \nabla \mathcal{F}(\mathbf{X}_m)$

\Return{$\mathbf{X}_m$}
\end{algorithm}

Next, we sample unlabeled locations to initialize the inducing points $\mathbf{X}_m$ of the SGP used to solve the SP problem. As per the findings of \cite{BurtRW20}, the initial inducing points' locations significantly affect the quality of the sparse approximation that can be obtained with SGPs. We recommend using methods such as the conditional variance~\citep{BurtRW20} or K-means~\citep{Bishop06} on the randomly sampled training dataset to initialize the inducing points.

We use the sampled unlabeled training set and the inducing points to initialize an SGP. Here, we use the spare variational free energy-based Gaussian process~(SVGP) as it is the SGP counterpart to our formulation. However, as mentioned in the previous subsection, any SGP approach can be used in our algorithm to address the SP problem.

We then optimize the inducing point locations of the SGP via the complexity and trace terms of the lower bound $\mathcal{F}$; the data fit term is disabled by the zero labeled training set. Since the lower bound $\mathcal{F}$ is fully differentiable, we can use gradient-based approaches to efficiently optimize the inducing points in continuous spaces. This gives us the optimized inducing points, which, in turn, represent the solution sensor placement locations.

\subsection{Greedy-SGP: Greedy Discrete Space Solutions}
\label{greedy-sgp}

Now consider the case when we want to limit the solution of the SP problem to a discrete set of candidate locations, either a subset of the training points or any other arbitrary set of points. In this case, we can generalize the inducing points selection approach for SGPs outlined in~\cite{Titsias09} to handle non-differentiable data domains.

\begin{algorithm}[h]
\caption{Greedy-SGP approach for obtaining sensor placements in discrete environments (i.e., sensor placements limited to a given set of candidate sensor locations) using a greedy selection approach. Here, $\theta$ are the hyperparameters learned from either historical data or expert knowledge, $\mathcal{S}$ is the set of candidate sensor placement locations, $\Phi$ is a random distribution defined within the boundaries of the environment $\mathcal{V}$, $s$ is the number of required sensors, and $n$ is the number of random unlabeled locations used to train the SGP.}
\label{alg:Greedy-SGP}
\SetKwInput{kwLoop}{Loop until}
\KwIn{$\theta, \mathcal{V}, \mathcal{S}, \Phi$, $s$, $n$}
\KwOut{Sensor placements $\mathcal{A} \subset \mathcal{S}$, where $|\mathcal{A}| = s$}

$\mathbf{X} \sim \Phi(\mathcal{V})$ \ \text{/ / \tenrm{Draw $n$ unlabeled locations}}

\tcc{Initialize the SGP with zero mean and zero labeled data}
 
$\varphi = \mathcal{SGP}(\text{mean}=0, \theta; \mathbf{X}, \mathbf{y}=\mathbf{0})$

\tcc{Sequentially select each of the solution inducing point locations using the criteria defined in terms of the optimization bound $\mathcal{F}$ of the SGP $\varphi$}

$\mathbf{X}_m = \{ \emptyset \}$

\Repeat{$|\mathbf{X}_m| = s$}{
$\mathbf{x}_m^* =  \argmax_{\mathbf{x} \in \mathcal{S} \backslash \mathbf{X}_m} \mathcal{F}(\mathbf{X}_m \cup \{\mathbf{x}\}) - \mathcal{F}(\mathbf{X}_m)$ \\

$\mathbf{X}_m \leftarrow \mathbf{X}_m \cup \mathbf{x}_m^*$
}

\Return{$\mathbf{X}_m$}
\end{algorithm}

The approach~(Algorithm~\ref{alg:Greedy-SGP}) entails sequentially selecting the inducing points $\mathbf{X}_m$ from the candidate set $\mathcal{S}$ using a greedy approach (Equation~\ref{eq:vfe_delta}). It considers the increment in the SVGP's optimization bound $\mathcal{F}$ as the maximization criteria. In each iteration, we select the point $\mathbf{x}$ that results in the largest increment in the SVGP's bound $\mathcal{F}$ upon being added to the current inducing points set $\mathbf{X}_m$:

\begin{equation}
\label{eq:vfe_delta}
\mathbf{X}_m \leftarrow \mathbf{X}_m \cup \{ \argmax_{\mathbf{x} \in \mathcal{S} \backslash \mathbf{X}_m} \mathcal{F}(\mathbf{X}_m \cup \{\mathbf{x}\}) - \mathcal{F}(\mathbf{X}_m) \}\,. \\
\end{equation}

Here $\mathbf{X}_m$ is the set of inducing points/sensing locations, and $\mathcal{S} \backslash \mathbf{X}_m$ is the set of remaining candidate locations after excluding the current inducing points set $\mathbf{X}_m$.

\textbf{Comparison with Mutual Information:} The Greedy-SGP approach has a few interesting similarities to the mutual information~(MI) based sensor placement approach by \cite{krauseSG08}. The MI approach uses a full GP to evaluate MI between the sensing locations and the rest of the environment to be monitored. The MI-based criteria shown below was used to greedily select sensing locations:

\begin{equation}
MI(\mathbf{X}_m \cup \{\mathbf{x}\}) - MI(\mathbf{X}_m) = H(\mathbf{x}|\mathbf{X}_m) - H(\mathbf{x}|\mathcal{S} \backslash \mathbf{X}_m)\,,
\end{equation}

where $\mathbf{X}_m$ is the set of selected sensing locations, and $\mathcal{S} \backslash \mathbf{X}_m$ is the set of all candidate locations in the environment excluding the current sensor locations $\mathbf{X}_m$. A Gaussian process (GP) with known kernel parameters was used to evaluate the entropy terms. The SVGP's optimization bound-based selection criterion to obtain discrete solutions using the greedy algorithm is equivalent to maximizing the following:

\begin{equation}
\label{eq:vfe_delta_greedy}
\Delta\mathcal{F} = \text{KL}(q(f_i|\mathbf{f}_m)||p{(f_i|\mathbf{y})}) - \text{KL}(p(f_i|\mathbf{f}_m)||p{(f_i|\mathbf{y})})\,.
\end{equation}

The first KL term measures the divergence between the sparse distribution $q$ over $f_i$ (the latent variable corresponding to $\mathbf{x}$) given the latents of the inducing points set $\mathbf{X}_m$, and the exact conditional given the training set labels $\mathbf{y}$ (the conditional uses the training set inputs $\mathbf{X}$ as well). The second term acts as a normalization term that measures the divergence between the exact conditional over $f_i$ given the latents of the inducing points and the same given the training labels. Please refer to Appendix~\ref{sec:theory} for the derivation of Equation~\ref{eq:vfe_delta_greedy}.

A key difference between the Greedy-SGP and the MI approach is that we use efficient cross-entropy (in the KL terms) to account for the whole environment. In contrast, the MI approach uses the computationally expensive entropy term $H(\mathbf{x}|\mathcal{S} \backslash \mathbf{X}_m)$. However, the overall formulation of both approaches is similar. We validate this empirically in the experiments section. 

\subsection{Discrete-SGP: Gradient-based Discrete Space Solutions}
\label{disc-sgp}

The problem with any greedy selection algorithm is its inherent sequential selection procedure. Given any sequentially optimized solution, it may be possible to improve the solution by replacing one or more of the solution locations with different locations from the candidate set. Thus, selecting the sensing locations sequentially restricts the solution placements to a subset of the solution space. We could find a better solution by simultaneously optimizing all the sensing locations as such an approach would account for their combined effect instead of considering only the incremental effect of the solution locations in each selection iteration.

\begin{algorithm}[ht]
\caption{Discrete-SGP approach for obtaining sensor placements in discrete environments (i.e., sensor placements limited to a given set of candidate sensor locations) using gradient descent. Here, $\theta$ are the hyperparameters learned from either historical data or expert knowledge, $\mathcal{S}$ is the set of candidate sensor placement locations, $\Phi$ is a random distribution defined within the boundaries of the environment $\mathcal{V}$, $s$ is the number of required sensors, $n$ is the number of random unlabeled locations used to train the SGP, and $\gamma$ is the SGP learning rate.}
\label{alg:Discrete-SGP}
\SetKwInput{kwLoop}{Loop until}
\KwIn{$\theta, \mathcal{V}, \mathcal{S}, \Phi$, $s$, $n$, $\gamma$}
\KwOut{Sensor placements $\mathcal{A} \subset \mathcal{S}$, where $|\mathcal{A}| = s$}

\tcc{Get the $s$ continuous space sensor placements using the gradient based approach Continuous-SGP~(Algorithm~\ref{alg:Continuous-SGP})}
$\mathbf{X}_m = \text{Continuous-SGP}(\theta, \mathcal{V}, \Phi, s, n, \gamma)$

\tcp{Compute pairwise $L_2$ distances}
$\mathbf{C} = \mathbf{0}^{|\mathbf{X}_m| \times |\mathcal{S}|}$

\For{$i\leftarrow 0$ \KwTo $|\mathbf{X}_m|$}{
    \For{$j\leftarrow 0$ \KwTo $|\mathcal{S}|$}{
        $\mathbf{C}[i][j] \leftarrow  ||\mathbf{X}_m[i]-\mathcal{S}[j]||_2$
    }
}

\tcc{Solve the assignment problem $\mathcal{H}$ to assign the $s$ continuous space inducing points $\mathbf{X}_m$ to locations in the candidate set $\mathcal{S}$}
$A = \mathcal{H}(\mathbf{C})$

\tcc{Use the assignments $A$ to index the candidate set $\mathcal{S}$ and get the discrete space solution}
$\mathbf{X}_m^* = \mathcal{S}[A]$

\Return{$\mathbf{X}_m^*$}
\end{algorithm}

Our approach (Algorithm~\ref{alg:Discrete-SGP}) to this problem is to simultaneously optimize all the inducing points in the continuous input space using gradient descent and then map the solution to the discrete candidate solution space $\mathcal{S}$. We can map the continuous space solutions to discrete sets by treating the mapping problem as an assignment problem~\citep{burkardDM12}, i.e., as a weighted bipartite matching problem. The assignment problem requires one to find the minimal cost matching of a set of items to another set of items given their pairwise costs. We compute the pairwise Euclidean distances between the continuous space inducing points and the discrete space candidate set locations $\mathcal{S}$. The distances are then used as the costs in an assignment problem. One could even use covariances that are appropriately transformed, instead of distances, in the mapping operation to account for the correlations in the environment.

The solution of the assignment problem gives us points in the discrete candidate set closest to the continuous space solution set. Such a solution could be superior to the greedy solution since the points in the continuous space solution set are simultaneously optimized using gradient descent instead of being sequentially selected. Although the gradient-based solution could get stuck in a local optimum, in our experiments, we found that the gradient-based discrete solutions are on par or better than the greedy solutions while being substantially faster to optimize.

\section{Experiments}

\begin{figure*}[!ht]
   \centering
    \begin{subfigure}{0.22\textwidth}
        \includegraphics[width=\textwidth]{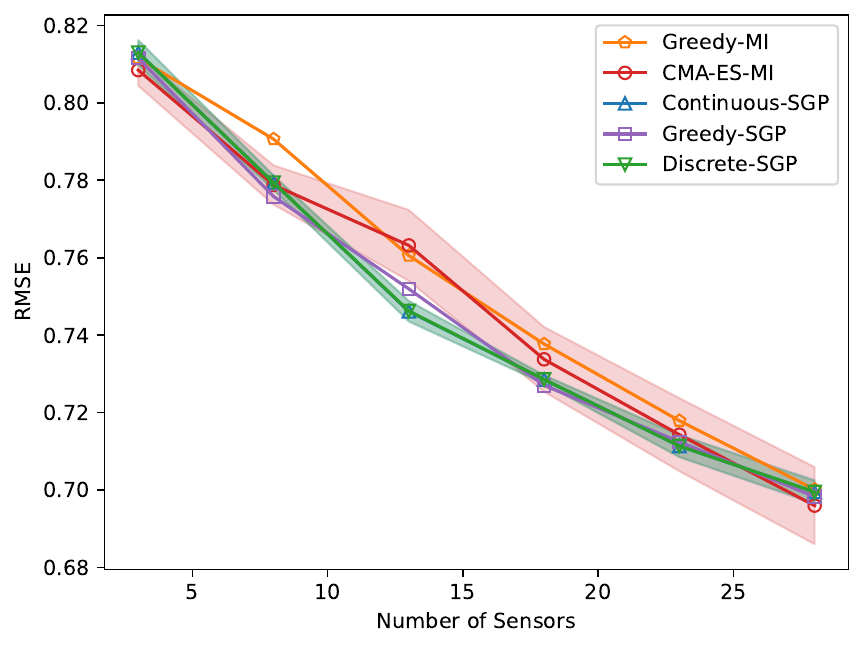}
        \caption{Temperature dataset}
        \label{fig:rmse-intel}
    \end{subfigure}
    \hfill
    \begin{subfigure}{0.22\textwidth}
        \includegraphics[width=\textwidth]{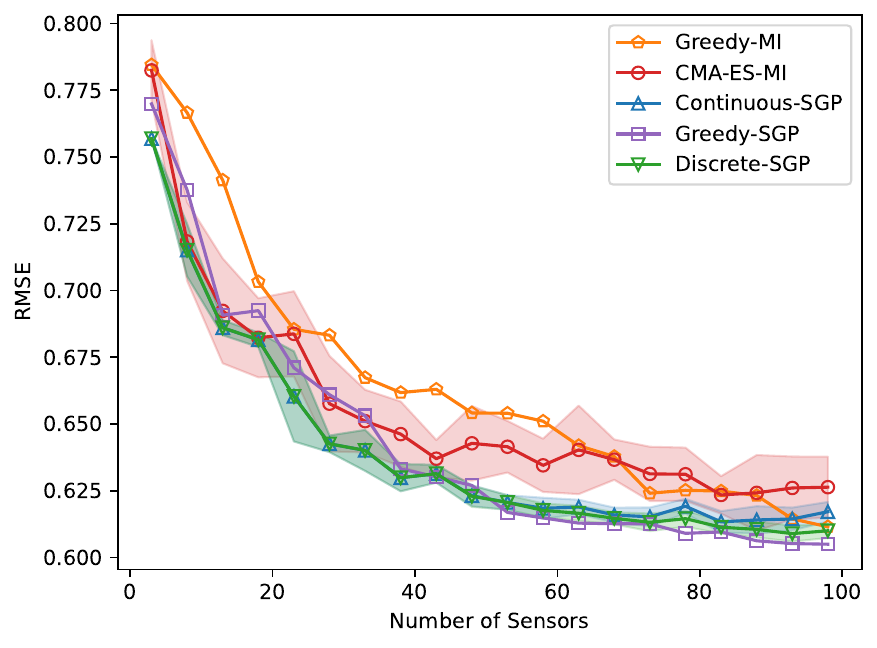}
        \caption{Precipitation dataset}
        \label{fig:rmse-precip}
    \end{subfigure}
    \hfill
    \begin{subfigure}{0.22\textwidth}
        \includegraphics[width=\textwidth]{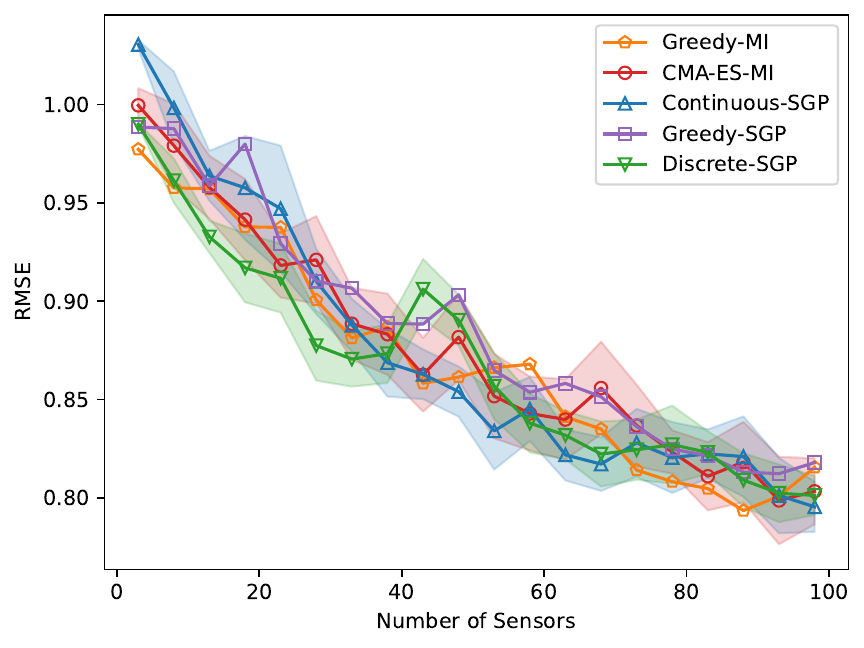}
        \caption{Soil dataset}
        \label{fig:rmse-soil}
    \end{subfigure}
    \hfill
    \begin{subfigure}{0.22\textwidth}
        \includegraphics[width=\textwidth]{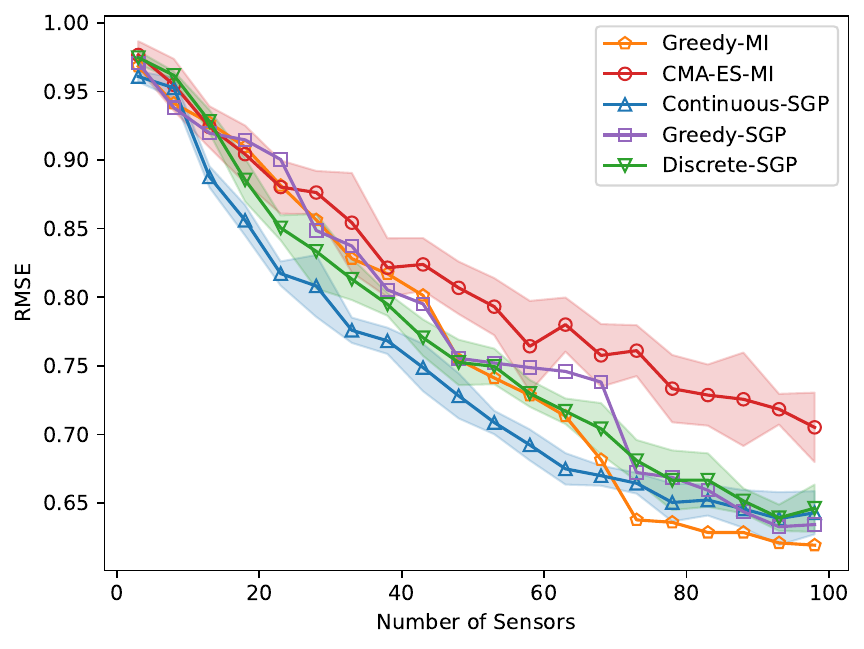}
        \caption{Salinity dataset}
        \label{fig:rmse-sal}
    \end{subfigure}
    \caption{The mean and standard deviation of the RMSE vs number of sensors for the Intel, precipitation, soil, and salinity datasets (lower is better).}
    \label{fig:rmse-benchmark}
\end{figure*}

\begin{figure*}[!ht]
   \centering
    \begin{subfigure}{0.22\textwidth}
        \includegraphics[width=\textwidth]{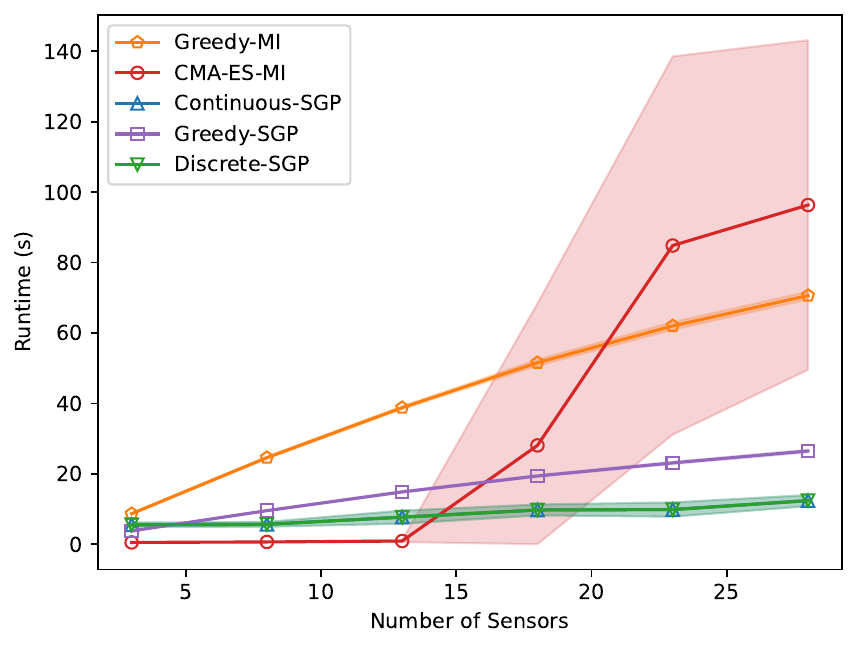}
        \caption{Temperature dataset}
        \label{fig:time-intel}
    \end{subfigure}
    \hfill
    \begin{subfigure}{0.22\textwidth}
        \includegraphics[width=\textwidth]{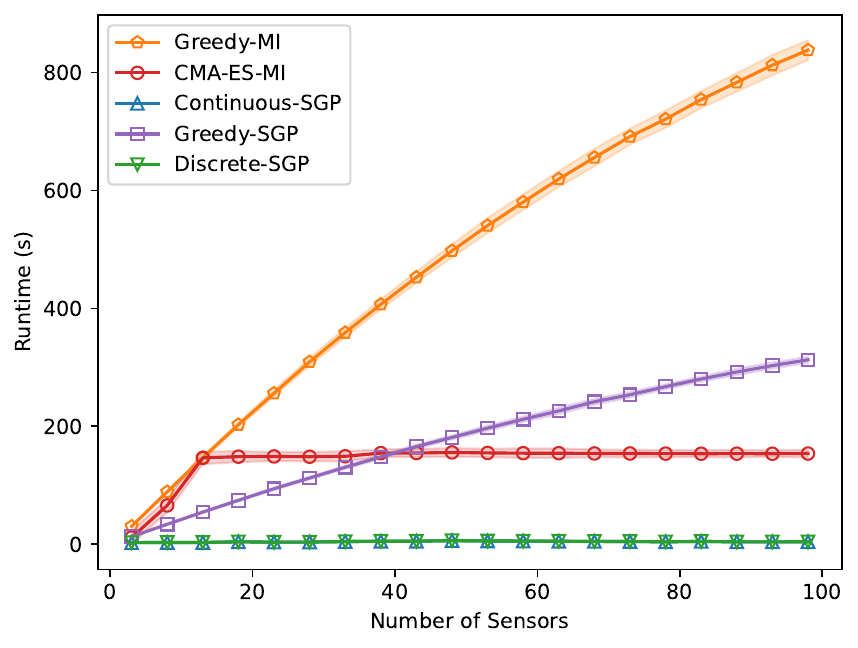}
        \caption{Precipitation dataset}
        \label{fig:time-precip}
    \end{subfigure}
    \hfill
    \begin{subfigure}{0.22\textwidth}
        \includegraphics[width=\textwidth]{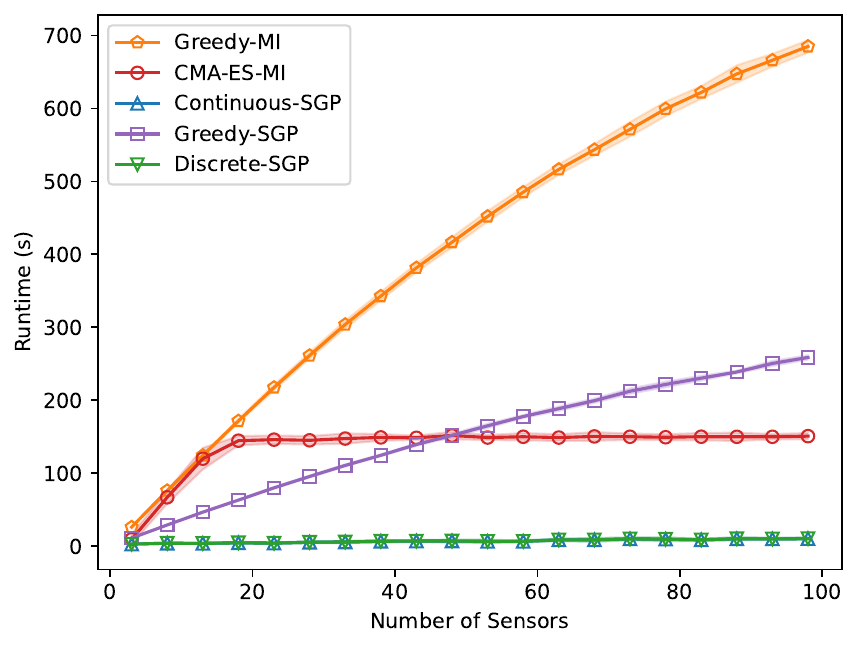}
        \caption{Soil dataset}
        \label{fig:time-soil}
    \end{subfigure}
    \hfill
    \begin{subfigure}{0.22\textwidth}
        \includegraphics[width=\textwidth]{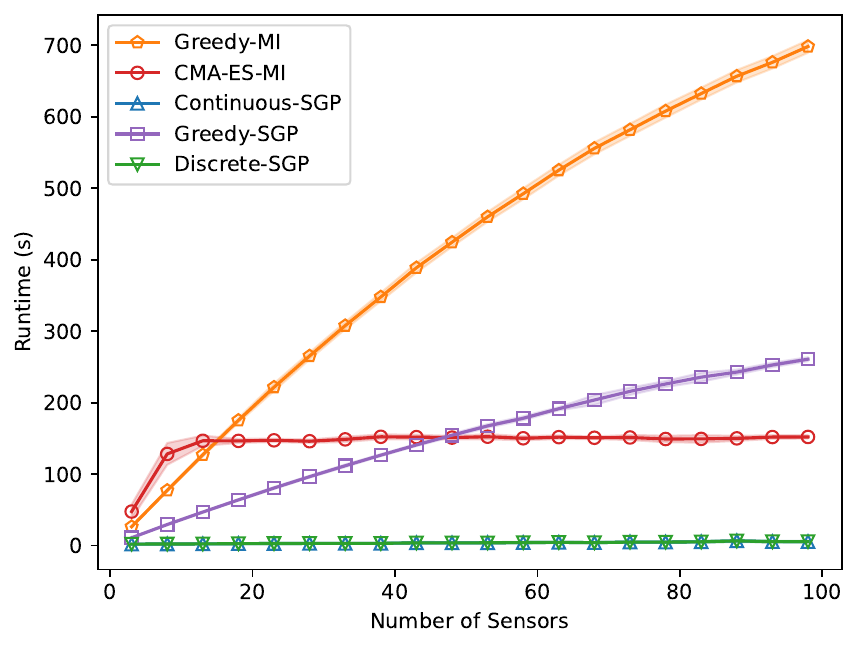}
        \caption{Salinity dataset}
        \label{fig:time-sal}
    \end{subfigure}
    \caption{The mean and standard deviation of the Runtime vs number of sensors for the Intel, precipitation, soil, and salinity datasets (lower is better).}
    \label{fig:time-benchmark}
\end{figure*}

\begin{figure*}[!ht]
   \centering
    \begin{subfigure}{0.22\textwidth}
        \includegraphics[width=\textwidth]{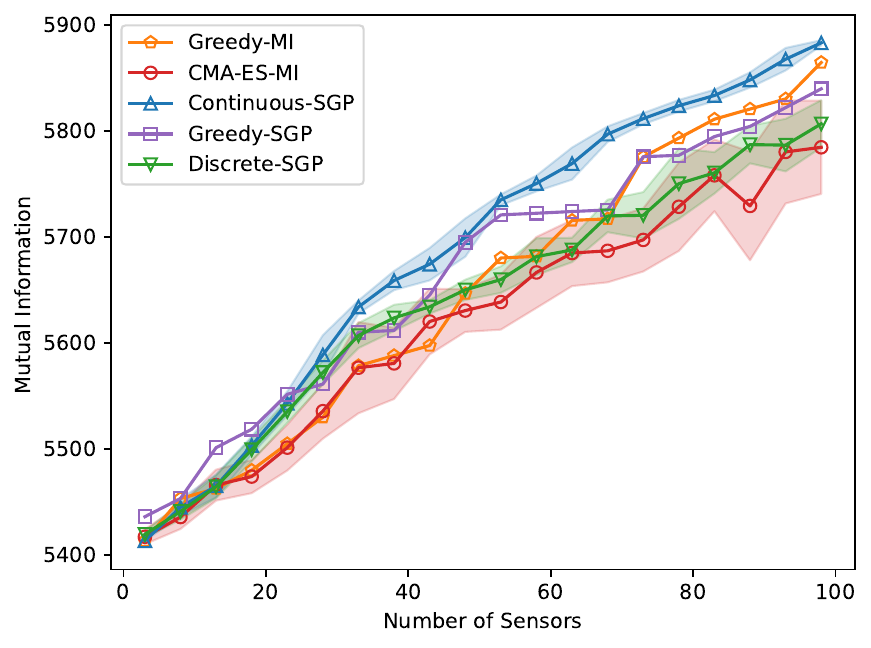}
        \caption{Soil dataset MI}
        \label{fig:mi-soil}
    \end{subfigure}
    \hfill
    \begin{subfigure}{0.22\textwidth}
        \includegraphics[width=\textwidth]{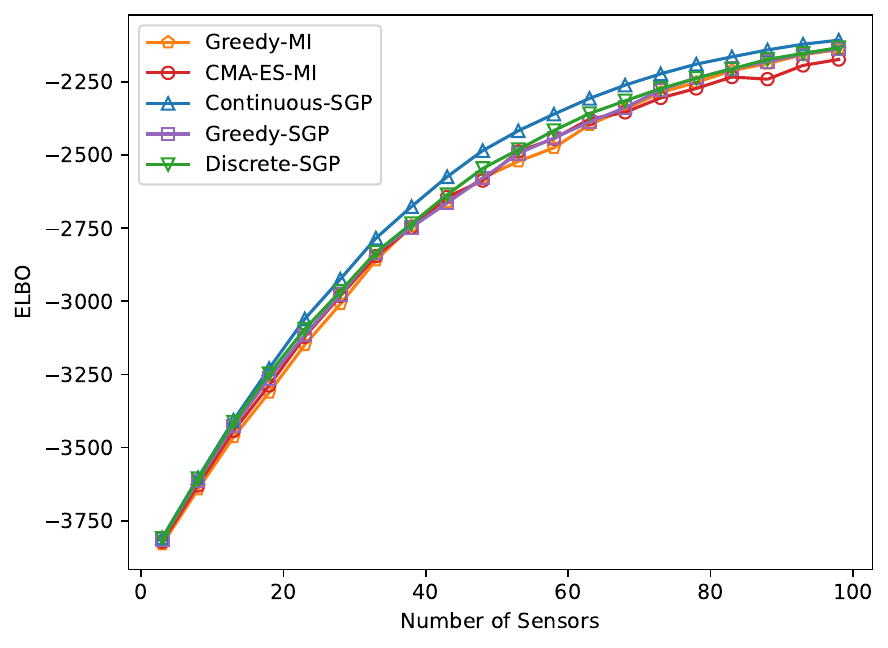}
        \caption{Soil dataset ELBO}
        \label{fig:elbo-soil}
    \end{subfigure}
    \hfill
    \begin{subfigure}{0.22\textwidth}
        \includegraphics[width=\textwidth]{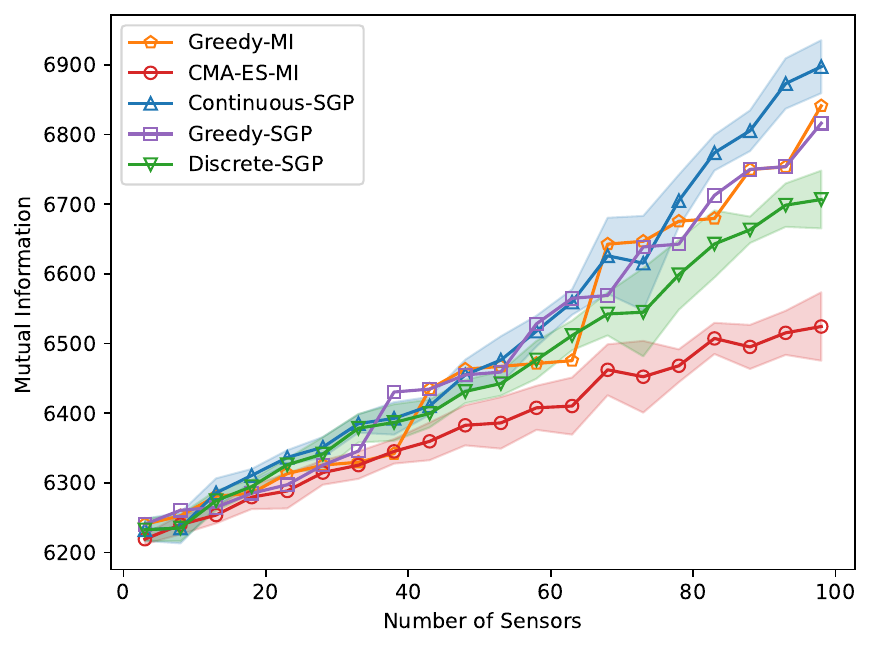}
        \caption{Salinity dataset MI}
        \label{fig:mi-sal}
    \end{subfigure}
    \hfill
    \begin{subfigure}{0.22\textwidth}
        \includegraphics[width=\textwidth]{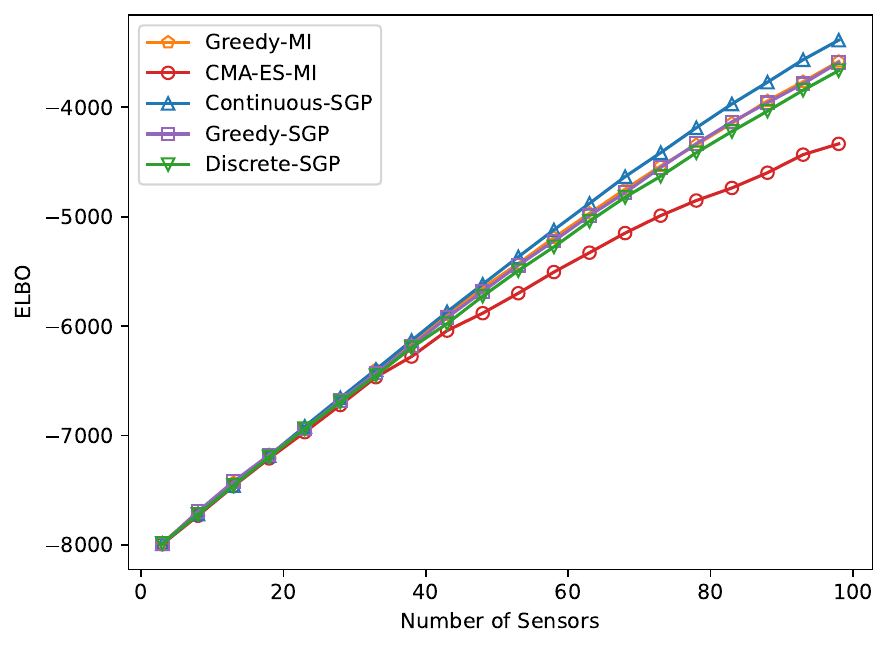}
        \caption{Salinity dataset ELBO}
        \label{fig:elbo-sal}
    \end{subfigure}
    \caption{Comparison of the MI and SVGP's lower bound (ELBO) for the soil and salinity datasets. The mean and standard deviation of the MI vs number of sensors (a), (c)  and SVGP's lower bound (ELBO) vs number of sensors (b), (d).}
    \label{fig:mi-benchmark}
\end{figure*}

We demonstrate our methods on four datasets—Intel lab temperature~\citep{bodikHGMPT04}, precipitation~\citep{brethertonWDWB99}, soil moisture~\citep{SoilData}, and ROMS ocean salinity~\citep{Shchepetkin05}. The datasets are representative of real-world sensor placement problems and some of these have been previously used as benchmarks~\citep{krauseSG08}. We used an RBF kernel~\citep{RasmussenW05} in these experiments. 

The Intel lab temperature dataset contains indoor temperature data collected from 54 sensors deployed in the Intel Berkeley Research lab.  The precipitation dataset contains daily precipitation data from 167 sensors around Oregon, U.S.A, in 1994. The US soil moisture dataset contains moisture readings from the continental USA, and the ROMS dataset contains salinity data from the Southern California Bight region. We uniformly sampled 150 candidate sensor placement locations in the soil and salinity datasets. 

For each dataset, we used a small portion of the data to learn the kernel parameters. and used a Gaussian process~(GP) to reconstruct the data field in the environment from each method's solution placements. The GP was initialized with the learned kernel function, and the solution sensing locations and their corresponding ground truth labels were used as the training set in the GP. We evaluated our data field reconstructions using the root-mean-square error (RMSE).

We benchmarked our approaches—Continuous-SGP~(Section \ref{Continuous-SGP}), Greedy-SGP~(Section \ref{greedy-sgp}), and Discrete-SGP~(Section \ref{disc-sgp}). We also evaluated the performance of the approach presented in \citet{krauseSG08}, which maximizes mutual information (MI) using the greedy algorithm~(Greedy-MI) in discrete environments. Additionally, we utilized the approach outlined in \citet{HitzGGPS17}, which employs a covariance matrix adaptation evolution strategy to maximize MI~(CMA-ES-MI) as another baseline, given its capability to handle continuous environments. We chose these baselines as they are closely related approaches that address the sensor placement problem by maximizing MI computed using GPs and have demonstrated strong performance on existing benchmarks.

\begin{figure*}[!ht]
   \centering
    \begin{subfigure}{0.18\textwidth}
        \includegraphics[width=\textwidth]{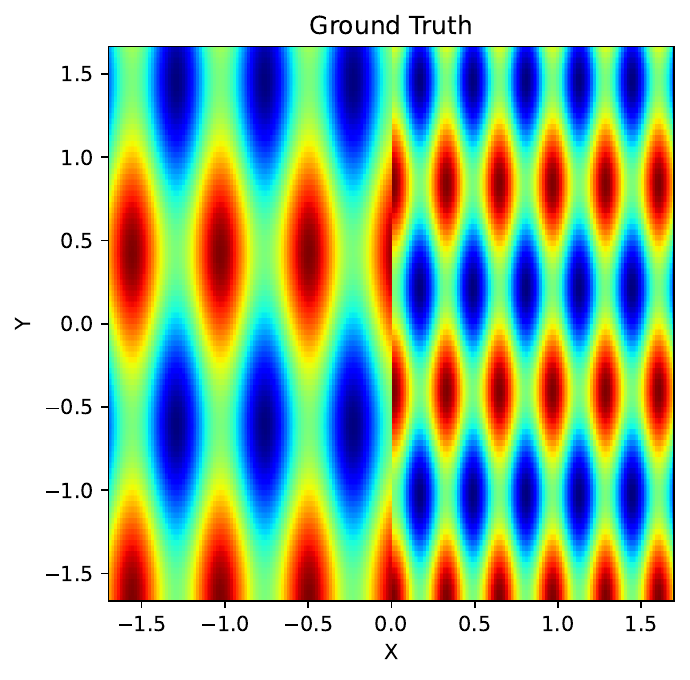}
        \subcaption{}
        \end{subfigure}
    \hfill
    \begin{subfigure}{0.18\textwidth}
        \includegraphics[width=\textwidth]{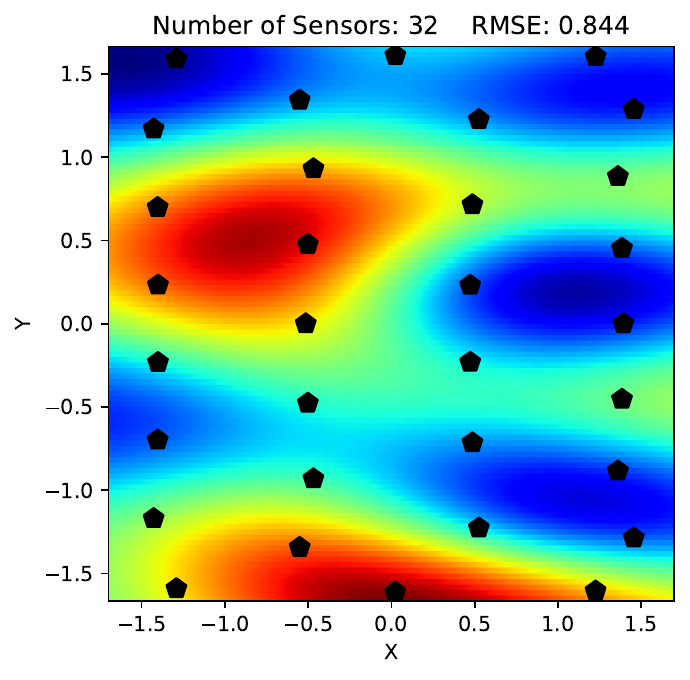}
        \subcaption{}
    \end{subfigure}
    \hfill
    \begin{subfigure}{0.18\textwidth}
        \includegraphics[width=\textwidth]{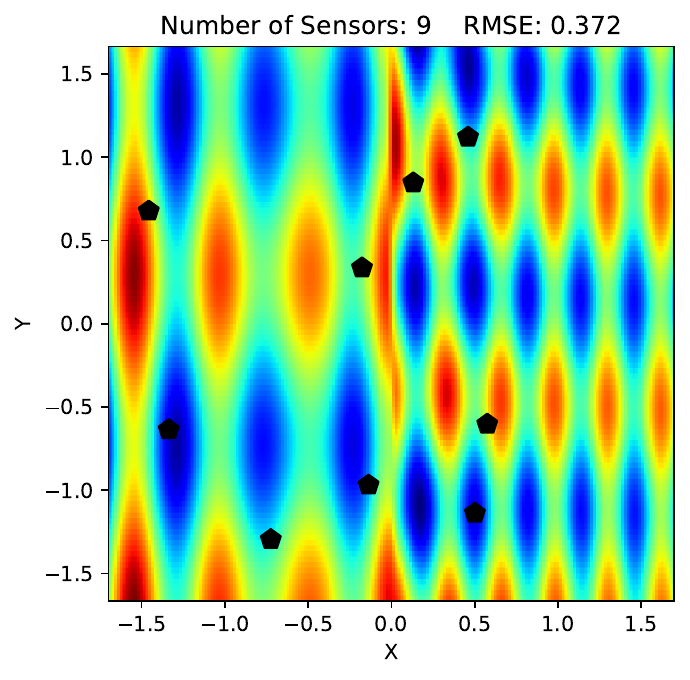}
        \subcaption{}
    \end{subfigure}
    \hfill
    \begin{subfigure}{0.18\textwidth}
        \includegraphics[width=\textwidth]{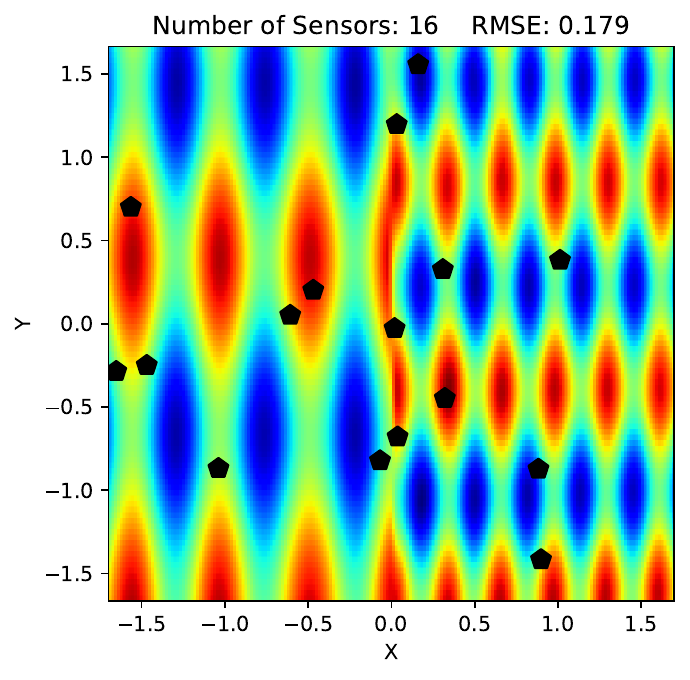}
        \subcaption{}
    \end{subfigure}
    \caption{A non-stationary environment. (a) Ground truth. Reconstructions from the Continuous-SGP solutions for (b) 32 sensing locations with a stationary RBF kernel, and (c) 9 and (d) 16 sensing locations with the neural kernel. The black pentagons represent the solution placements.}
    \label{fig:non-stat-benchmark}
\vspace{-5mm}
\end{figure*}

We computed the solution sensor placements for 3 to 100 sensors (in increments of 5) for all the datasets, except for the Intel dataset, which was tested for up to 30 placement locations since it has only 54 placement locations in total. The experiments were repeated 10 times and we report the mean and standard deviation of the RMSE and runtime results in Figures~\ref{fig:rmse-benchmark} and~\ref{fig:time-benchmark}. We see that our approaches' RMSE results are consistently on par or better that the baseline Greedy-MI and CMA-ES-MI approaches. 

Moreover, our approaches are substantially faster than the baseline approaches. Our Continuous-SGP approach is up to $6$ times faster than the baselines in the temperature dataset, up to $50$ times faster in the precipitation dataset, and up to $43$ times faster in the soil and salinity datasets. Our SGP-based approaches select sensing locations that  reduce the RMSE by maximizing the SGP's ELBO, which requires inverting only an $m \times m$ covariance matrix ($m \ll |\mathcal{S}|$, where $|\mathcal{S}|$ is the number of candidate locations). In contrast, both the baselines maximize MI, which requires inverting up to an $|\mathcal{S}| \times |\mathcal{S}|$ covariance matrix to place each sensor, which takes $\mathcal{O}(|\mathcal{S}|^3)$ time. As such, the computational cost difference is further exacerbated in the precipitation, soil, and salinity datasets, which have three times as many candidate locations as the temperature data. 

Also, the Continuous-SGP method consistently generates high quality results in our experiments; which is consistent with the findings of~\citet{BauerWR16}, who showed that SVGP's are able to recover the full GP posterior in regression tasks. Solving the assignment problem in our Discrete-SGP approach to map the Continuous-SGP solution to the discrete candidate set incurs a one-time $\mathcal{O}(m^3)$ computation cost that is negligible. Therefore our gradient-based approaches—Continuous-SGP and Discrete-SGP—converge at almost the same rate. Yet the Discrete-SGP retains the solution quality of the Continuous-SGP. 

The labeled data locations in the soil and salinity datasets used to train our kernel function were not aligned with our candidate locations for the discrete approaches. We chose this setup to demonstrate that we can learn the kernel parameters even if the data is not aligned with the candidate locations, or is from a different environment altogether.

In Figure~\ref{fig:mi-benchmark} we show the MI between the solution placements and the environments (2500 uniformly sampled locations) and the SVGP lower bound (ELBO). The label dependent data fit term in the ELBO was disabled to remove the data bias. Although the ELBO values differ from MI, they closely approximate the relative trends of the MI values. This supports our claim that the SVGP's ELBO behaves similarly to MI while being significantly cheaper to compute.

When using stationary kernels, our approach and approaches that maximize MI generate solution placements that are equally spaced. However, the solution placements are more informative when using non-stationary kernels. The following experiment demonstrates our approach in a non-stationary environment. We used a neural kernel~\citep{RemesHK18} to learn the non-stationary correlations in the environment (Figure~\ref{fig:non-stat-benchmark}). We see that our placements from the SGP approach with the neural kernel are not uniformly distributed and are able to achieve near-perfect reconstructions of the environment. Please refer to the Supplementary for further details on the above experiments and additional experiments on spatiotemporal sensor placement, sensor placement in an environment with obstacles, and comparison against two additional baselines.
\section{Conclusion}
We addressed the sensor placement problem for monitoring correlated data. We formulated the problem using variational approximation and showed that training SGPs on unlabeled data gives us ideal sensor placements in continuous spaces, thereby opening up the vast GP and SGP literature to the sensor placement problem and its variants involving constraints, non-point sensors, etc. The method also enables us to efficiently handle sensor placement in 3D spaces, spatio-temporally correlated spaces, and leverage the convergence rate proofs of SGPs. Furthermore, we presented an approach that uses the assignment problem to map the continuous domain solutions to discrete domains efficiently, giving us computationally efficient discrete solutions compared to the greedy approach. Our experiments on four real-world datasets demonstrated that our approaches result in both mutual information and reconstruction quality being on par or better than the existing approaches while substantially reducing the computation time. 

A key advantage of our approach is its differentiability with respect to the sensing locations, which allows us to incorporate it into deep neural networks optimized for more complex downstream tasks. We leverage this differentiability property in concurrent work to generalize our sensor placement approach to robotic informative path planning. Since our approach, and more generally GP-based approaches, rely on accurate kernel function parameters, we aim to develop online approaches to address this in our future work.

\comment{
\section*{Acknowledgements}
This work was funded in part by the UNC Charlotte Office of Research and Economic Development and by NSF under Award Number IIP-1919233.}

\bibliography{references}

\begin{thebibliography}{52}
\providecommand{\natexlab}[1]{#1}
\providecommand{\url}[1]{\texttt{#1}}
\expandafter\ifx\csname urlstyle\endcsname\relax
  \providecommand{\doi}[1]{doi: #1}\else
  \providecommand{\doi}{doi: \begingroup \urlstyle{rm}\Url}\fi

\bibitem[Bai et~al.(2006)Bai, Kumar, Xuan, Yun, and Lai]{BaiKXYL06}
Xiaole Bai, Santosh Kumar, Dong Xuan, Ziqiu Yun, and Ten~H. Lai.
\newblock {Deploying Wireless Sensors to Achieve Both Coverage and
  Connectivity}.
\newblock In \emph{Proceedings of the 7th ACM International Symposium on Mobile
  Ad Hoc Networking and Computing}, page 131–142, New York, NY, USA, 2006.

\bibitem[Bauer et~al.(2016)Bauer, van~der Wilk, and Rasmussen]{BauerWR16}
Matthias Bauer, Mark van~der Wilk, and Carl~Edward Rasmussen.
\newblock {Understanding Probabilistic Sparse Gaussian Process Approximations}.
\newblock In \emph{Advances in Neural Information Processing Systems}, page
  1533–1541, Red Hook, NY, USA, 2016.

\bibitem[Bigoni et~al.(2020)Bigoni, Zhang, and Hesthaven]{BigoniZH20}
Caterina Bigoni, Zhenying Zhang, and Jan~S. Hesthaven.
\newblock Systematic sensor placement for structural anomaly detection in the
  absence of damaged states.
\newblock \emph{Computer Methods in Applied Mechanics and Engineering},
  371:\penalty0 113315, 2020.
\newblock ISSN 0045-7825.

\bibitem[Bishop(2006)]{Bishop06}
Christopher Bishop.
\newblock \emph{Pattern Recognition and Machine Learning}.
\newblock Springer, New York, 2006.

\bibitem[Blei et~al.(2017)Blei, Kucukelbir, and McAuliffe]{BleiKM17}
David~M. Blei, Alp Kucukelbir, and Jon~D. McAuliffe.
\newblock Variational inference: A review for statisticians.
\newblock \emph{Journal of the American Statistical Association}, 112\penalty0
  (518):\penalty0 859–877, April 2017.

\bibitem[Bodik et~al.(2004)Bodik, Hong, Guestrin, Madden, Paskin, and
  Thibaux]{bodikHGMPT04}
Peter Bodik, Wei Hong, Carlos Guestrin, Sam Madden, Mark Paskin, and Romain
  Thibaux.
\newblock Intel lab data.
\newblock \emph{Online dataset}, 2004.
\newblock URL \url{http://db.csail.mit.edu/labdata/labdata.html}.

\bibitem[Breitenmoser et~al.(2010)Breitenmoser, Schwager, Metzger, Siegwart,
  and Rus]{BreitenmoserSMSR10}
Andreas Breitenmoser, Mac Schwager, Jean-Claude Metzger, Roland Siegwart, and
  Daniela Rus.
\newblock Voronoi coverage of non-convex environments with a group of networked
  robots.
\newblock In \emph{2010 IEEE International Conference on Robotics and
  Automation}, pages 4982--4989, 2010.

\bibitem[Bretherton et~al.(1999)Bretherton, Widmann, Dymnikov, Wallace, and
  Blad{\'e}]{brethertonWDWB99}
Christopher Bretherton, Martin Widmann, Valentin Dymnikov, John Wallace, and
  Ileana Blad{\'e}.
\newblock The effective number of spatial degrees of freedom of a time-varying
  field.
\newblock \emph{Journal of Climate}, 12\penalty0 (7):\penalty0 1990--2009,
  1999.

\bibitem[Buisson-Fenet et~al.(2020)Buisson-Fenet, Solowjow, and
  Trimpe]{FenetST20}
Mona Buisson-Fenet, Friedrich Solowjow, and Sebastian Trimpe.
\newblock {Actively Learning Gaussian Process Dynamics}.
\newblock In Alexandre~M. Bayen, Ali Jadbabaie, George Pappas, Pablo~A.
  Parrilo, Benjamin Recht, Claire Tomlin, and Melanie Zeilinger, editors,
  \emph{Proceedings of the 2nd Conference on Learning for Dynamics and
  Control}, volume 120 of \emph{Proceedings of Machine Learning Research},
  pages 5--15. PMLR, 10--11 Jun 2020.

\bibitem[Burkard et~al.(2012)Burkard, Dell'Amico, and Martello]{burkardDM12}
Rainer Burkard, Mauro Dell'Amico, and Silvano Martello.
\newblock \emph{{Assignment Problems: revised reprint}}.
\newblock SIAM, Philadelphia, USA, 2012.

\bibitem[Burt et~al.(2019)Burt, Rasmussen, and Van Der~Wilk]{BurtRW19}
David Burt, Carl~Edward Rasmussen, and Mark Van Der~Wilk.
\newblock {Rates of Convergence for Sparse Variational {G}aussian Process
  Regression}.
\newblock In Kamalika Chaudhuri and Ruslan Salakhutdinov, editors,
  \emph{Proceedings of the 36th International Conference on Machine Learning},
  volume~97, pages 862--871. PMLR, Jun 2019.

\bibitem[Burt et~al.(2020)Burt, Rasmussen, and van~der Wilk]{BurtRW20}
David~R. Burt, Carl~Edward Rasmussen, and Mark van~der Wilk.
\newblock {Convergence of Sparse Variational Inference in Gaussian Processes
  Regression}.
\newblock \emph{Journal of Machine Learning Research}, 21\penalty0
  (131):\penalty0 1--63, 2020.

\bibitem[{Copernicus Climate Change Service}(2018)]{Copernicusccs18}
{Copernicus Climate Change Service}.
\newblock Ozone monthly gridded data from 1970 to present derived from
  satellite observations, 2018.

\bibitem[Cortes et~al.(2004)Cortes, Martinez, Karatas, and Bullo]{CortesMKB04}
Jorge Cortes, Sonia Martinez, Timur Karatas, and Francesco Bullo.
\newblock Coverage control for mobile sensing networks.
\newblock \emph{IEEE Transactions on Robotics and Automation}, 20\penalty0
  (2):\penalty0 243--255, 2004.

\bibitem[de~Berg et~al.(2008)de~Berg, Cheong, van Kreveld, and
  Overmars]{DeBergCKO08}
Mark de~Berg, Otfried Cheong, Marc van Kreveld, and Mark Overmars.
\newblock \emph{Computational Geometry: Algorithms and Applications}.
\newblock Springer-Verlag, Berlin, third edition, 2008.

\bibitem[Francis et~al.(2019)Francis, Ott, Marchant, and Ramos]{FrancisOMR19}
Gilad Francis, Lionel Ott, Roman Marchant, and Fabio Ramos.
\newblock Occupancy map building through {B}ayesian exploration.
\newblock \emph{The International Journal of Robotics Research}, 38\penalty0
  (7):\penalty0 769--792, 2019.
\newblock \doi{10.1177/0278364919846549}.

\bibitem[Goodfellow et~al.(2016)Goodfellow, Bengio, and
  Courville]{GoodfellowBC16}
I.~Goodfellow, Y.~Bengio, and A.~Courville.
\newblock \emph{Deep Learning}.
\newblock Adaptive Computation and Machine Learning series. MIT Press, 2016.
\newblock ISBN 9780262035613.

\bibitem[Hain(2013)]{SoilData}
Christopher Hain.
\newblock {NASA SPoRT-LiS Soil Moisture Products}, 2013.
\newblock URL
  \url{https://www.drought.gov/data-maps-tools/nasa-sport-lis-soil-moisture-products}.

\bibitem[Hamelijnck et~al.(2021)Hamelijnck, Wilkinson, Loppi, Solin, and
  Damoulas]{HamelijnckWLSD21}
Oliver Hamelijnck, William~J. Wilkinson, Niki~Andreas Loppi, Arno Solin, and
  Theo Damoulas.
\newblock {Spatio-Temporal Variational Gaussian Processes}.
\newblock In A.~Beygelzimer, Y.~Dauphin, P.~Liang, and J.~Wortman Vaughan,
  editors, \emph{Advances in Neural Information Processing Systems}, 2021.

\bibitem[Hensman et~al.(2013)Hensman, Fusi, and Lawrence]{HensmanFL13}
James Hensman, Nicol\`{o} Fusi, and Neil~D. Lawrence.
\newblock {Gaussian Processes for Big Data}.
\newblock In \emph{Proceedings of the Twenty-Ninth Conference on Uncertainty in
  Artificial Intelligence}, page 282–290, Arlington, Virginia, USA, 2013.
  AUAI Press.

\bibitem[Hitz et~al.(2017)Hitz, Galceran, Garneau, Pomerleau, and
  Siegwart]{HitzGGPS17}
Gregory Hitz, Enric Galceran, Marie-Eve Garneau, François Pomerleau, and
  Roland Siegwart.
\newblock {Adaptive Continuous-Space Informative Path Planning for Online
  Environmental Monitoring}.
\newblock \emph{Journal of Field Robotics}, 34\penalty0 (8):\penalty0
  1427--1449, 2017.

\bibitem[Husain and Caselton(1980)]{HusainC80}
T.~Husain and W.~F. Caselton.
\newblock {Hydrologic Network Design Methods and Shannon's Information Theory}.
\newblock \emph{IFAC Proceedings Volumes}, 13\penalty0 (3):\penalty0 259--267,
  1980.
\newblock {IFAC} Symposium on Water and Related Land Resource Systems,
  Cleveland, OH, USA, May 1980.

\bibitem[Jakkala and Akella(2022)]{JakkalaA22}
Kalvik Jakkala and Srinivas Akella.
\newblock {Probabilistic Gas Leak Rate Estimation Using Submodular Function
  Maximization With Routing Constraints}.
\newblock \emph{IEEE Robotics and Automation Letters}, 7\penalty0 (2):\penalty0
  5230--5237, 2022.

\bibitem[Kingma and Ba(2015)]{KingmaB14}
Diederik~P. Kingma and Jimmy Ba.
\newblock {Adam: A Method for Stochastic Optimization}.
\newblock In \emph{International Conference on Learning Representations
  (Poster)}, 2015.

\bibitem[Krause et~al.(2008)Krause, Singh, and Guestrin]{krauseSG08}
Andreas Krause, Ajit Singh, and Carlos Guestrin.
\newblock {Near-Optimal Sensor Placements in Gaussian Processes: Theory,
  Efficient Algorithms and Empirical Studies}.
\newblock \emph{Journal of Machine Learning Research}, 9\penalty0 (8):\penalty0
  235--284, 2008.

\bibitem[Li et~al.(2021)Li, Liu, and Wang]{LiLW21}
Bangjun Li, Haoran Liu, and Ruzhu Wang.
\newblock Efficient sensor placement for signal reconstruction based on
  recursive methods.
\newblock \emph{IEEE Transactions on Signal Processing}, 69:\penalty0
  1885--1898, 2021.
\newblock \doi{10.1109/TSP.2021.3063495}.

\bibitem[Lin et~al.(2019)Lin, Chowdhury, Wang, and Terejanu]{LinCWT19}
Xiao Lin, Asif Chowdhury, Xiaofan Wang, and Gabriel Terejanu.
\newblock Approximate computational approaches for {B}ayesian sensor placement
  in high dimensions.
\newblock \emph{Information Fusion}, 46:\penalty0 193--205, 2019.
\newblock ISSN 1566-2535.

\bibitem[Ma et~al.(2017)Ma, Liu, and Sukhatme]{MaLS17}
Kai-Chieh Ma, Lantao Liu, and Gaurav~S. Sukhatme.
\newblock {Informative Planning and Online Learning with Sparse Gaussian
  Processes}.
\newblock In \emph{2017 IEEE International Conference on Robotics and
  Automation (ICRA)}, pages 4292--4298, 2017.

\bibitem[McKay et~al.(1979)McKay, Beckman, and Conover]{MckayBC79}
M.~McKay, Richard Beckman, and William Conover.
\newblock {A Comparison of Three Methods for Selecting Values of Input
  Variables in the Analysis of Output From a Computer Code}.
\newblock \emph{Technometrics}, 21:\penalty0 239--245, 05 1979.

\bibitem[Murphy(2022)]{Murphy22}
Kevin~P. Murphy.
\newblock \emph{Probabilistic Machine Learning: An introduction}.
\newblock MIT Press, Massachusetts, 2022.

\bibitem[Nemhauser et~al.(1978)Nemhauser, Wolsey, and Fisher]{NemhauserWF78}
G.~L. Nemhauser, L.~A. Wolsey, and M.~L. Fisher.
\newblock {An analysis of approximations for maximizing submodular set
  functions-I}.
\newblock \emph{Mathematical Programming}, 14\penalty0 (1):\penalty0 265--294,
  Dec 1978.

\bibitem[Quinonero-Candela et~al.(2007)Quinonero-Candela, Rasmussen, and
  Williams]{CandelaRW07}
Joaquin Quinonero-Candela, Carl~Edward Rasmussen, and Christopher K.~I.
  Williams.
\newblock {A}pproximation {M}ethods for {G}aussian {P}rocess {R}egression.
\newblock In \emph{Large-Scale Kernel Machines}, pages 203--223. MIT Press,
  2007.

\bibitem[Ramsden(2009)]{Ramsden09}
Daryn Ramsden.
\newblock \emph{Optimization approaches to sensor placement problems}.
\newblock PhD thesis, Department of Mathematical Sciences, Rensselaer
  Polytechnic Institute, August 2009.

\bibitem[Rasmussen and Williams(2005)]{RasmussenW05}
Carl~Edward Rasmussen and Christopher K.~I. Williams.
\newblock \emph{{Gaussian Processes for Machine Learning}}.
\newblock MIT Press, Cambridge, USA, 2005.

\bibitem[Remes et~al.(2018)Remes, Heinonen, and Kaski]{RemesHK18}
Sami Remes, Markus Heinonen, and Samuel Kaski.
\newblock Neural non-stationary spectral kernel.
\newblock \emph{ArXiv}, 2018.

\bibitem[Sadeghi et~al.(2022)Sadeghi, Asghar, and Smith]{SadeghiAS22}
Armin Sadeghi, Ahmad~B. Asghar, and Stephen~L. Smith.
\newblock Distributed multi-robot coverage control of non-convex environments
  with guarantees.
\newblock \emph{IEEE Transactions on Control of Network Systems}, pages 1--12,
  2022.

\bibitem[Salam and Hsieh(2019)]{SalamH19}
Tahiya Salam and M.~Ani Hsieh.
\newblock Adaptive sampling and reduced-order modeling of dynamic processes by
  robot teams.
\newblock \emph{IEEE Robotics and Automation Letters}, 4\penalty0 (2):\penalty0
  477--484, 2019.

\bibitem[Schreiber et~al.(2020)Schreiber, Bilmes, and Noble]{SchreiberBN20}
Jacob Schreiber, Jeffrey Bilmes, and William~Stafford Noble.
\newblock apricot: {S}ubmodular selection for data summarization in {P}ython.
\newblock \emph{Journal of Machine Learning Research}, 21\penalty0
  (161):\penalty0 1--6, 2020.

\bibitem[Schwager et~al.(2017)Schwager, Vitus, Powers, Rus, and
  Tomlin]{SchwagerVPRT17}
Mac Schwager, Michael~P. Vitus, Samantha Powers, Daniela Rus, and Claire~J.
  Tomlin.
\newblock Robust adaptive coverage control for robotic sensor networks.
\newblock \emph{IEEE Transactions on Control of Network Systems}, 4\penalty0
  (3):\penalty0 462--476, 2017.

\bibitem[Semaan(2017)]{Semaan17}
R.~Semaan.
\newblock Optimal sensor placement using machine learning.
\newblock \emph{Computers \& Fluids}, 159:\penalty0 167--176, 2017.
\newblock ISSN 0045-7930.

\bibitem[Shchepetkin and McWilliams(2005)]{Shchepetkin05}
Alexander~F. Shchepetkin and James~C. McWilliams.
\newblock {The regional oceanic modeling system (ROMS): a split-explicit,
  free-surface, topography-following-coordinate oceanic model}.
\newblock \emph{Ocean Modelling}, 9\penalty0 (4):\penalty0 347--404, 2005.
\newblock ISSN 1463-5003.

\bibitem[Shewry and Wynn(1987)]{ShewryW87}
M.~C. Shewry and H.~P. Wynn.
\newblock Maximum entropy sampling.
\newblock \emph{Journal of Applied Statistics}, 14\penalty0 (2):\penalty0
  165--170, 1987.

\bibitem[Snelson and Ghahramani(2006)]{SnelsonG06}
Edward Snelson and Zoubin Ghahramani.
\newblock {Sparse Gaussian Processes using Pseudo-inputs}.
\newblock In Y.~Weiss, B.~Sch\"{o}lkopf, and J.~Platt, editors, \emph{Advances
  in Neural Information Processing Systems}, volume~18. MIT Press, 2006.

\bibitem[Suryan and Tokekar(2020)]{SuryanT20}
Varun Suryan and Pratap Tokekar.
\newblock {Learning a Spatial Field in Minimum Time With a Team of Robots}.
\newblock \emph{IEEE Transactions on Robotics}, 36\penalty0 (5):\penalty0
  1562--1576, 2020.

\bibitem[Tajnafoi et~al.(2021)Tajnafoi, Arcucci, Mottet, Vouriot,
  Molina-Solana, Pain, and Guo]{TajnafoiAMVMPG21}
Gabor Tajnafoi, Rossella Arcucci, Laetitia Mottet, Carolanne Vouriot, Miguel
  Molina-Solana, Christopher Pain, and Yi-Ke Guo.
\newblock Variational gaussian process for optimal sensor placement.
\newblock \emph{Applications of Mathematics}, 66\penalty0 (2):\penalty0
  287--317, Apr 2021.
\newblock ISSN 1572-9109.

\bibitem[Titsias(2009)]{Titsias09}
Michalis Titsias.
\newblock {Variational Learning of Inducing Variables in Sparse Gaussian
  Processes}.
\newblock In David van Dyk and Max Welling, editors, \emph{Proceedings of the
  Twelth International Conference on Artificial Intelligence and Statistics},
  pages 567--574, Florida, USA, 2009. PMLR.

\bibitem[{van der Wilk} et~al.(2020){van der Wilk}, Dutordoir, John, Artemev,
  Adam, and Hensman]{WilkDJAAH20}
Mark {van der Wilk}, Vincent Dutordoir, ST~John, Artem Artemev, Vincent Adam,
  and James Hensman.
\newblock {A Framework for Interdomain and Multioutput Gaussian Processes}.
\newblock \emph{ArXiv}, 2020.

\bibitem[Wang et~al.(2020)Wang, Li, and Chen]{WangLC20}
Zhi Wang, Han-Xiong Li, and Chunlin Chen.
\newblock Reinforcement learning-based optimal sensor placement for
  spatiotemporal modeling.
\newblock \emph{IEEE Transactions on Cybernetics}, 50\penalty0 (6):\penalty0
  2861--2871, 2020.

\bibitem[Whitman et~al.(2021)Whitman, Maske, Kingravi, and
  Chowdhary]{WhitmanMKC21}
Joshua Whitman, Harshal Maske, Hassan~A. Kingravi, and Girish Chowdhary.
\newblock {Evolving Gaussian Processes and Kernel Observers for Learning and
  Control in Spatiotemporally Varying Domains: With Applications in
  Agriculture, Weather Monitoring, and Fluid Dynamics}.
\newblock \emph{IEEE Control Systems}, 41:\penalty0 30--69, 2021.

\bibitem[Wilkinson et~al.(2021)Wilkinson, S\"arkk\"a, and Solin]{wilkinsonSS21}
William~J. Wilkinson, Simo S\"arkk\"a, and Arno Solin.
\newblock {{B}ayes-{N}ewton Methods for Approximate {B}ayesian Inference with
  {PSD} Guarantees}.
\newblock \emph{CoRR}, abs/2111.01721, 2021.

\bibitem[Wu and Zidek(1992)]{WuZ92}
Shiying Wu and James~V. Zidek.
\newblock {An entropy-based analysis of data from selected NADP/NTN network
  sites for 1983–1986}.
\newblock \emph{Atmospheric Environment. Part A. General Topics}, 26\penalty0
  (11):\penalty0 2089--2103, 1992.

\bibitem[Zhu et~al.(2021)Zhu, Chung, Lawrance, Siegwart, and
  Alonso-Mora]{ZhuCLSA21}
Hai Zhu, Jen~Jen Chung, Nicholas~R.J. Lawrance, Roland Siegwart, and Javier
  Alonso-Mora.
\newblock {Online Informative Path Planning for Active Information Gathering of
  a 3D Surface}.
\newblock In \emph{2021 IEEE International Conference on Robotics and
  Automation (ICRA)}, pages 1488--1494, 2021.

\end{thebibliography}

\newpage

\onecolumn

\title{Efficient Sensor Placement from Regression with Sparse Gaussian Processes\\(Supplementary Material)}
\maketitle

\appendix

\tableofcontents

\clearpage
\section{Theory}
\label{sec:theory}

This section shows the derivation of the SVGP evidence lower bound's delta term, which is used to contrast it with the MI approach's delta term~\citep{krauseSG08}, and details why the SGVP's evidence lower bound is not submodular. 

\subsection{Preliminary}

\subsubsection{Properties of Entropy}

\begin{enumerate}
    \item Joint entropy can be decomposed into the sum of conditional entropy and marginal entropy~\citep{Bishop06}:
    $$
    \begin{aligned}
    H(X, Y) =& H(X|Y) + H(Y) \\
            =& H(Y|X) + H(X)\,.
    \end{aligned}
    $$
    
    \item The reverse KL divergence is the cross entropy minus entropy~\citep{Murphy22}:
    $$
    \text{KL}(q||p) = H_p(q) - H(q)\,.
    $$
\end{enumerate}

\subsubsection{Submodularity}
\hfill\\
A set function $f$ is submodular if it has the following diminishing returns property for sets $X, Y,$ and $T$, with $u$ being an element of the set $T$ that is not already in $Y$~\citep{NemhauserWF78}:

\begin{eqnarray*}
f(X \cup \{u\}) - f(X) \geq f(Y \cup \{u\}) - f(Y) \\
\forall X \subseteq Y \subset T \ \ \text{and} \ \ u \in T \backslash Y\,.
\end{eqnarray*}

\subsection{SVGP Evidence Lower Bound's Delta Term Expansion}

The lower bound of the SVGP~\citep{Titsias09} is given by:

\begin{equation}
\mathcal{F} = \frac{n}{2} \log(2\pi) + \frac{1}{2} \log |\mathbf{Q}_{nn} + \sigma_{\text{noise}}^2 I| + \frac{1}{2} \mathbf{y}^\top (\mathbf{Q}_{nn} + \sigma_{\text{noise}}^2 I)^{-1} \mathbf{y} - \frac{1}{2\sigma_{\text{noise}}^2} Tr(\mathbf{K}_{nn} - \mathbf{Q}_{nn})\,,
\end{equation}
where $\mathbf{K}_{nn}$ is the covariance matrix computed using the SGP's kernel function on the $n$ training samples $\mathbf{X}$, $\mathbf{Q}_{nn} = \mathbf{K}_{nm} \mathbf{K}_{mm}^{-1} \mathbf{K}_{mn}$, the subscript $m$ corresponds to the inducing points set, $\sigma_{\text{noise}}$ is the noise variance, and $\mathbf{y}$ is the vector containing the training set labels.

Assume that the inducing points are a subset of the training set indexed by $m \subset \{ 1, ..., n\}$. Let ($\mathbf{X}_m, \mathbf{f}_m$) be the set of inducing points locations and their corresponding latent variables. Similarly, let ($\mathbf{X}, \mathbf{f}$) be the training set locations and latent variables. Here $n$ is the index set corresponding to the training dataset, and $m$ is the index set corresponding to the inducing points. Note that we use the same notation as~\citet{Titsias09}, who also used $n$ and $m$ to  denote the cardinality of these sets. We know that the SVGP evidence lower bound can be written as follows~\citep{Bishop06} for inducing points $\mathbf{X}_m$:

\begin{equation}
\begin{aligned}
\mathcal{F}(\mathbf{X}_m) =& -\text{KL}(q_m(\mathbf{f})||p(\mathbf{f}|\mathbf{y})) + \log p(\mathbf{y}) \\
=& - H_{p{(\mathbf{f}|\mathbf{y})}}(q_m(\mathbf{f})) + H(q_m(\mathbf{f})) + \log p(\mathbf{y})\,.
\end{aligned}
\end{equation}

Here $q_m$ is the variational distribution of the SGP with the $m$ inducing points. We index the $n$ training set points excluding the $m$ inducing set points as the set difference $n-m$. We can use the above to formulate the increments (delta term) in the SVGP lower bound upon adding a new inducing point $\mathbf{x}_i$ such that $i \in n-m$ as follows:

\begin{equation}
\label{step1}
\begin{aligned}
\Delta\mathcal{F}(\mathbf{X}_m, \{ \mathbf{x}_i \}) =& \mathcal{F}(\mathbf{X}_m \cup \{ \mathbf{x}_i \}) - \mathcal{F}(\mathbf{X}_m) \\
=& -\text{KL}(q_{m+1}(\mathbf{f}) || p(\mathbf{f}|\mathbf{y})) + \text{KL}(q_{m}(\mathbf{f}) || p(\mathbf{f}|\mathbf{y})) \\
=& - H_{p{(\mathbf{f}|\mathbf{y})}}(q_{m+1}(\mathbf{f})) + H(q_{m+1}(\mathbf{f})) + H_{p{(\mathbf{f}|\mathbf{y})}}(q_m(\mathbf{f})) - H(q_m(\mathbf{f})) \\
=& \underbrace{(H(q_{m+1}(\mathbf{f})) - H(q_m(\mathbf{f})))}_{\Delta h_1} - \underbrace{(H_{p{(\mathbf{f}|\mathbf{y})}}(q_{m+1}(\mathbf{f})) - H_{p{(\mathbf{f}|\mathbf{y})}}(q_m(\mathbf{f})))}_{\Delta h_2} \,.
\end{aligned}
\end{equation}

The last equation above is similar to the KL divergence, except that each entropy term here $\Delta h_j$ is the difference of two entropies. We can use the following expansion of the variational distribution $q_m$ to simplify the above:

\begin{equation}
\begin{aligned}
q_m(\mathbf{f}) &= p(\mathbf{f}_{n-(m+1)}, f_i | \mathbf{f}_m) \phi(\mathbf{f}_m) \\
&= p(\mathbf{f}_{n-(m+1)} | f_i, \mathbf{f}_m) p(f_i|\mathbf{f}_m) \phi(\mathbf{f}_m) \,.
\end{aligned}
\end{equation}

Here we factorized the variational distribution over $\mathbf{f}$ as the product of the variational distribution $\phi$ over the latents $\mathbf{f}_m$ parametrized with the $m$ inducing points $\mathbf{X}_m$ and the conditional distribution $p$ over the remaining data points $n-m$ computed using conditioning; $f_i$ corresponds to the additional data sample added to the $m$ inducing points. Similar to the above we can expand $q_{m+1}(\mathbf{f})$ as follows:

\begin{equation}
\begin{aligned}
q_{m+1}(\mathbf{f}) &= p(\mathbf{f}_{n-(m+1)}| \mathbf{f}_{m+1}) \phi(\mathbf{f}_{m+1}) \\
&= p(\mathbf{f}_{n-(m+1)} |\mathbf{f}_m, f_i) \phi(f_i|\mathbf{f}_m) \phi(\mathbf{f}_m) \,.
\end{aligned}
\end{equation}

Instead of using the conditional $p(f_i|\mathbf{f}_m)$ as we did for $q_m(\mathbf{f})$, here the distribution over $f_i$ is from the variational distribution $\phi(f_i|\mathbf{f}_m)$. Since all the inducing points are explicitly given in the variational distribution, the joint variational distribution over the inducing points can be decomposed as the product of marginals. We can now plug the decomposed variational distribution back into Equation~\ref{step1} to get the following using the chain rule of entropy:

\begin{equation}
\begin{aligned}
\Delta h_1 =& H(q_{m+1}(\mathbf{f})) - H(q_m(\mathbf{f})) \\
=& H(p(\mathbf{f}_{n-(m+1)} | f_i, \mathbf{f}_m) \phi(f_i|\mathbf{f}_m) \phi(\mathbf{f}_m)) - H(p(\mathbf{f}_{n-(m+1)} | f_i, \mathbf{f}_m) p(f_i|\mathbf{f}_m) \phi(\mathbf{f}_m)) \\
=& \cancel{H(p(\mathbf{f}_{n-(m+1)} | f_i, \mathbf{f}_m))} + H(\phi(f_i|\mathbf{f}_m)) + \cancel{H(\phi(\mathbf{f}_m))} - \cancel{H(p(\mathbf{f}_{n-(m+1)} | f_i, \mathbf{f}_m))} - H(p(f_i|\mathbf{f}_m)) - \cancel{H(\phi(\mathbf{f}_m))} \\
=& H(\phi(f_i|\mathbf{f}_m)) - H(p(f_i|\mathbf{f}_m)) \,.
\end{aligned}
\end{equation}

Similar to the above, we can get $\Delta h_2 = H_{p{(f_i|\mathbf{y})}}(\phi(f_i|\mathbf{f}_m)) - H_{p{(f_i|\mathbf{y})}}(p(f_i|\mathbf{f}_m))$. This gives us the following:

\begin{equation}
\begin{aligned}
\Delta\mathcal{F}(\mathbf{X}_m, \{ \mathbf{x}_i \}) &= H(\phi(f_i|\mathbf{f}_m)) - H(p(f_i|\mathbf{f}_m)) - H_{p{(f_i|\mathbf{y})}}(\phi(f_i|\mathbf{f}_m)) + H_{p{(f_i|\mathbf{y})}}(p(f_i|\mathbf{f}_m)) \\
&= (H(\phi(f_i|\mathbf{f}_m)) - H_{p{(f_i|\mathbf{y})}}(\phi(f_i|\mathbf{f}_m))) - (H(p(f_i|\mathbf{f}_m)) - H_{p{(f_i|\mathbf{y})}}(p(f_i|\mathbf{f}_m))) \\
&= \text{KL}(\phi(f_i|\mathbf{f}_m)||p{(f_i|\mathbf{y})}) - \text{KL}(p(f_i|\mathbf{f}_m)||p{(f_i|\mathbf{y})}) \,.
\end{aligned}
\end{equation}

Consider $\mathbf{X}_m \subseteq \mathbf{X}_l \subset \mathbf{X}$ and $\mathbf{x}_i \in \mathbf{X} \backslash \mathbf{X}_l$:

\begin{equation}
\begin{aligned}
\Delta\mathcal{F}(\mathbf{X}_m, \{ \mathbf{x}_i \}) - \Delta\mathcal{F}(\mathbf{X}_l, \{ \mathbf{x}_i \}) \geq 0 \\
\text{KL}(\phi(f_i|\mathbf{f}_m)||p{(f_i|\mathbf{y})}) - \text{KL}(p(f_i|\mathbf{f}_m)||p{(f_i|\mathbf{y})}) - \text{KL}(\phi(f_i|\mathbf{f}_l)||p{(f_i|\mathbf{y})}) + \text{KL}(p(f_i|\mathbf{f}_l)||p{(f_i|\mathbf{y})}) \geq 0 \\
\end{aligned}
\end{equation}

One needs to show that the last equation above is true for the SVGP's lower bound to be submodular, which is not necessarily true in all cases. If we constrain the variational distribution $\phi$ to have a diagonal covariance matrix, the diagonal covariance matrix assumption allows us to drop the variational distribution's dependence on $f_m$ in $\Delta h_1$ and $\Delta h_2$. This gives us the following: 

\begin{equation}
\begin{aligned}
\Delta h_1 &= H(\phi(f_i)) - H(p(f_i|\mathbf{f}_m)) \\
\Delta h_2 &= H_{p{(f_i|\mathbf{y})}}(\phi(f_i)) - H_{p{(f_i|\mathbf{y})}}(p(f_i|\mathbf{f}_m)) \,.
\end{aligned}
\end{equation}

Now if we again consider $\mathbf{X}_m \subseteq \mathbf{X}_l \subset \mathbf{X}$ and $\mathbf{x}_i \in \mathbf{X} \backslash \mathbf{X}_l$:
\begin{equation}
\begin{aligned}
\Delta\mathcal{F}(\mathbf{X}_m, \{ \mathbf{x}_i \}) - \Delta\mathcal{F}(\mathbf{X}_l, \{ \mathbf{x}_i \}) \geq 0 \\
\text{KL}(\phi(f_i)||p{(f_i|\mathbf{y})}) - \text{KL}(p(f_i|\mathbf{f}_m)||p{(f_i|\mathbf{y})}) - \text{KL}(\phi(f_i)||p{(f_i|\mathbf{y})}) + \text{KL}(p(f_i|\mathbf{f}_l)||p{(f_i|\mathbf{y})}) \geq 0 \\
\cancel{H(\phi(f_i))} - H(p(f_i|\mathbf{f}_m)) - \cancel{H_{p{(f_i|\mathbf{y})}}(\phi(f_i))} + H_{p{(f_i|\mathbf{y})}}(p(f_i|\mathbf{f}_m)) - \ \ \ \ \ \ \ \ \\
\cancel{H(\phi(f_i))} + H(p(f_i|\mathbf{f}_l)) + \cancel{H_{p{(f_i|\mathbf{y})}}(\phi(f_i))} - H_{p{(f_i|\mathbf{y})}}(p(f_i|\mathbf{f}_l))  \geq 0 \\
H_{p{(f_i|\mathbf{y})}}(p(f_i|\mathbf{f}_m)) - H(p(f_i|\mathbf{f}_m)) - H_{p{(f_i|\mathbf{y})}}(p(f_i|\mathbf{f}_l)) + H(p(f_i|\mathbf{f}_l)) \geq 0 \\
\text{KL}(p(f_i|\mathbf{f}_m)||p{(f_i|\mathbf{y})}) - \text{KL}(p(f_i|\mathbf{f}_l)||p{(f_i|\mathbf{y})}) \geq 0 \\
\text{KL}(p(f_i|\mathbf{f}_m)||p{(f_i|\mathbf{y})}) \geq \text{KL}(p(f_i|\mathbf{f}_l)||p{(f_i|\mathbf{y})}) \\
\text{KL}(p(f_i|\mathbf{f}_m)||p{(f_i|\mathbf{y})}) \geq \text{KL}(p(f_i|\mathbf{f}_m \cup (\mathbf{f}_l \backslash \mathbf{f}_m))||p{(f_i|\mathbf{y})}) \\
\end{aligned}
\end{equation}

The last equation above is much simpler than the inequality obtained using a full covariance matrix in the variational distribution. Indeed all the distributions used in the KL divergence terms are Gaussian, and therefore have a closed form equation for the KL divergence. However, we still found the inequality far too complex to be able to conclusively prove that the SVGP's lower bound is submodular.

So we performed empirical tests to check for submodularity and found that the SVGP's ELBO with a full covariance matrix and a diagonal covariance matrix were not submodular. Nonetheless, we found that when using a diagonal covariance matrix, the bound was almost submodular. This finding is based on the quality of the solutions found by optimizing the lower bound using the naive greedy and lazy greedy algorithms. When using a diagonal covariance matrix, the solution inducing points' lower bounds from both the algorithms were very close. This suggests that even if one were to treat the lower bound as submodular and optimize it using the efficient lazy greedy algorithm, we would still get good solutions.

\clearpage
\section{Additional Experiments}
\label{add_exp}

\subsection{Spatiotemporal Sensor Placement}

We demonstrate our approach's scalability to large spatiotemporal data fields by finding placements for $500$ ozone concentration sensors across the planet. Note that the environment is the surface of a sphere in this example. We used a spatiotemporal-sparse variational Gaussian process~(ST-SVGP)~\citep{HamelijnckWLSD21} as it allows us to efficiently model spatiotemporal correlations in the data with time complexity linear in the number of time steps in the training set. We used Matern 3/2 kernels~\citep{RasmussenW05} to model the spatial and temporal correlations. All the model parameters were optimized with a learning rate of $0.01$, and the parameters were optimized using the Adam optimizer~\citep{KingmaB14}. The ST-SVGP was trained on the first six months of the monthly ozone data from 2018~\citep{Copernicusccs18}, and we used a subset of $1040$ uniformly distributed locations in the dataset as the training set and $100$ inducing points to learn the kernel parameters. The learned kernel function was then used in our sensor placement approach—Continuous-SGP (Algorithm~\ref{alg:Continuous-SGP})—to obtain the $500$ solution placements shown below. Note that the solution sensor placements are spatially fixed to monitor the spatiotemporal data. 

\begin{figure}[ht!]
    \centering
    \includegraphics[width=0.6\linewidth]{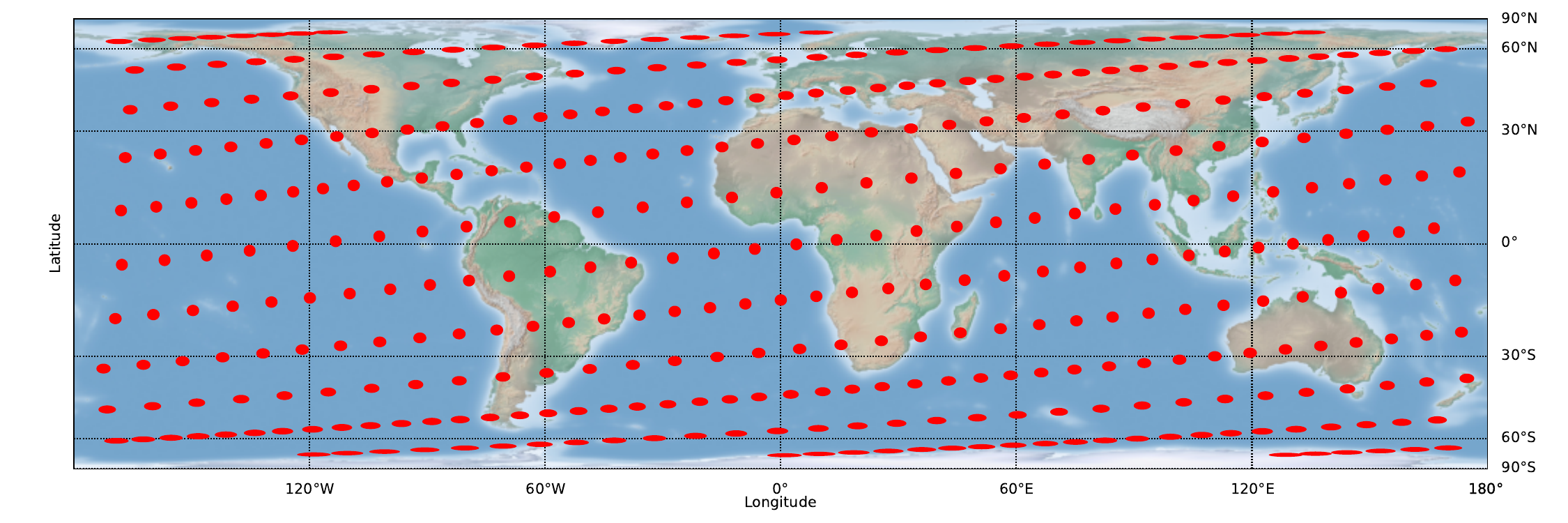}
    \caption{Placements for 500 sensors generated using the Continuous-SGP approach with an ST-SVGP. The red points are the sensor placements projected onto the 2D map using cylindrical equal-area projection.}
    \label{fig:st-placements}
\end{figure}

The solution placements are relatively uniformly distributed over the planet. This is because we used a stationary kernel function. However, in a real-world scenario, using a non-stationary kernel would give us even more informative sensing locations that can further leverage the non-stationary nature of the environment.

\subsection{Obstacle Avoidance}
\begin{figure}
    \centering
    \includegraphics[width=0.5\linewidth]{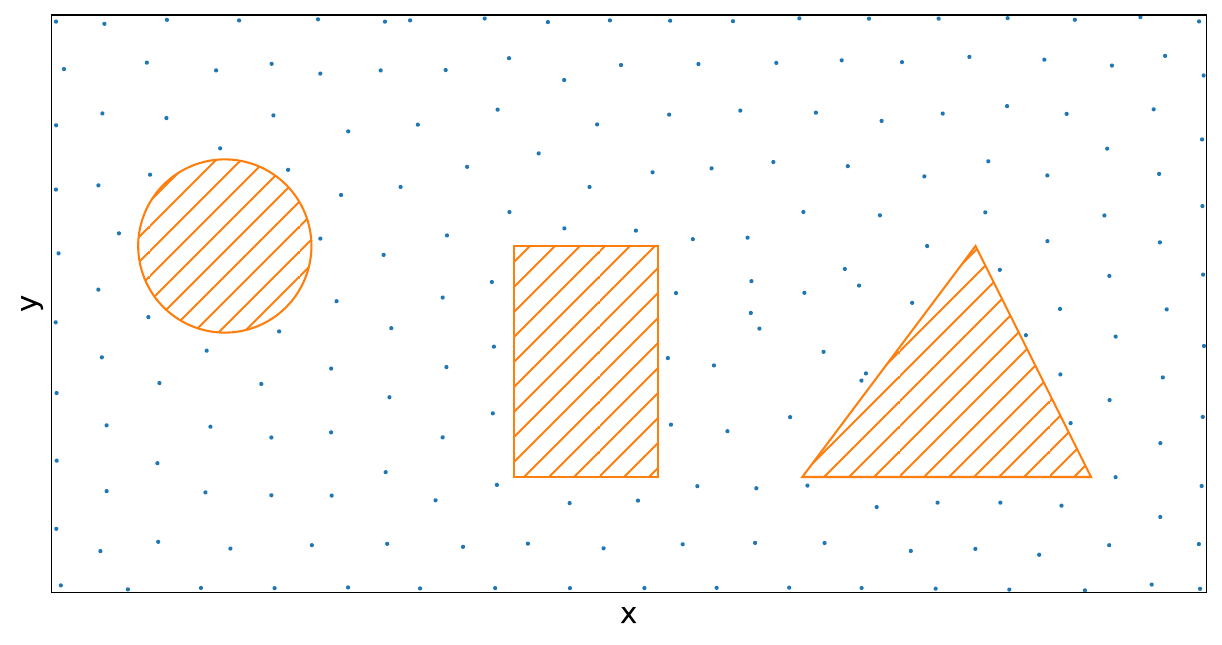}
    \caption{Placements for 200 sensors generated using the SGP approach. The hatched orange polygons represent obstacles in the environment and the blue points represent the solution sensor placements.}
    \label{fig:fov-placements}
\end{figure}

We handle obstacles in the environment by building an appropriate training dataset for the SGP. We remove the random samples in the SGP training set at locations in the interior of obstacles. Therefore the resulting training set has samples only in obstacle-free regions. Training an SGP on such data would result in inducing points that avoid the obstacles since placing the inducing points at locations with obstacles would not increase the likelihood of the training data used to optimize the SGP. Our obstacle avoidance approach is best suited for relatively large obstacles.

We now present our solution sensor placements in an environment with multiple obstacles (Figure~\ref{fig:fov-placements}). We trained the SGP using gradient descent on randomly sampled points in the environment where there were no obstacles and set all labels to zero. As we can see, the solution placements are well-spaced to ensure that the same information is not repeatedly collected. Also, our solution placements perfectly avoid the obstacles.

\subsection{Comparison with a Genetic Algorithm based Approach}

\citet{BigoniZH20} used the inducing points of the sparse Gaussian process similarly to our method. However, we identified that we can train SGPs for sensor placement in a \textit{sparsely supervised} manner by omitting the label-dependent data fit term from the lower bound (ELBO) of the SGP. This allows our method to be used even without a large set of labeled data from the environment, as required by \citep{BigoniZH20}. Moreover, \citet{BigoniZH20} utilized genetic algorithms to constrain the sensors within the bounds of the environment. However, this approach is computationally expensive and compromises the key advantage of SGPs—namely, their differentiability, which can be leveraged to address related problems, such as the informative path planning (IPP) problem. Our current paper introduced an efficient projection method based on the assignment problem, which enables us to employ standard gradient descent for the efficient optimization of sensor placements within the bounds of the environment.

Additionally, for completeness, the following experiment benchmarks our approach against \citet{BigoniZH20}. Since \citet{BigoniZH20} can only handle sensor placement in environments with labeled data, we modify their approach to handle unlabeled environments by using a kernel function with known parameters and disabling the data fit term in the SGP's optimization function. This leaves the optimization method as the key factor that affects the solution sensor placements; i.e., \citet{BigoniZH20} use genetic algorithms, and we use gradient descent to find the sensor placement.

\begin{figure}[!ht]
   \centering
    \begin{subfigure}{0.4\textwidth}
        \includegraphics[width=\textwidth]{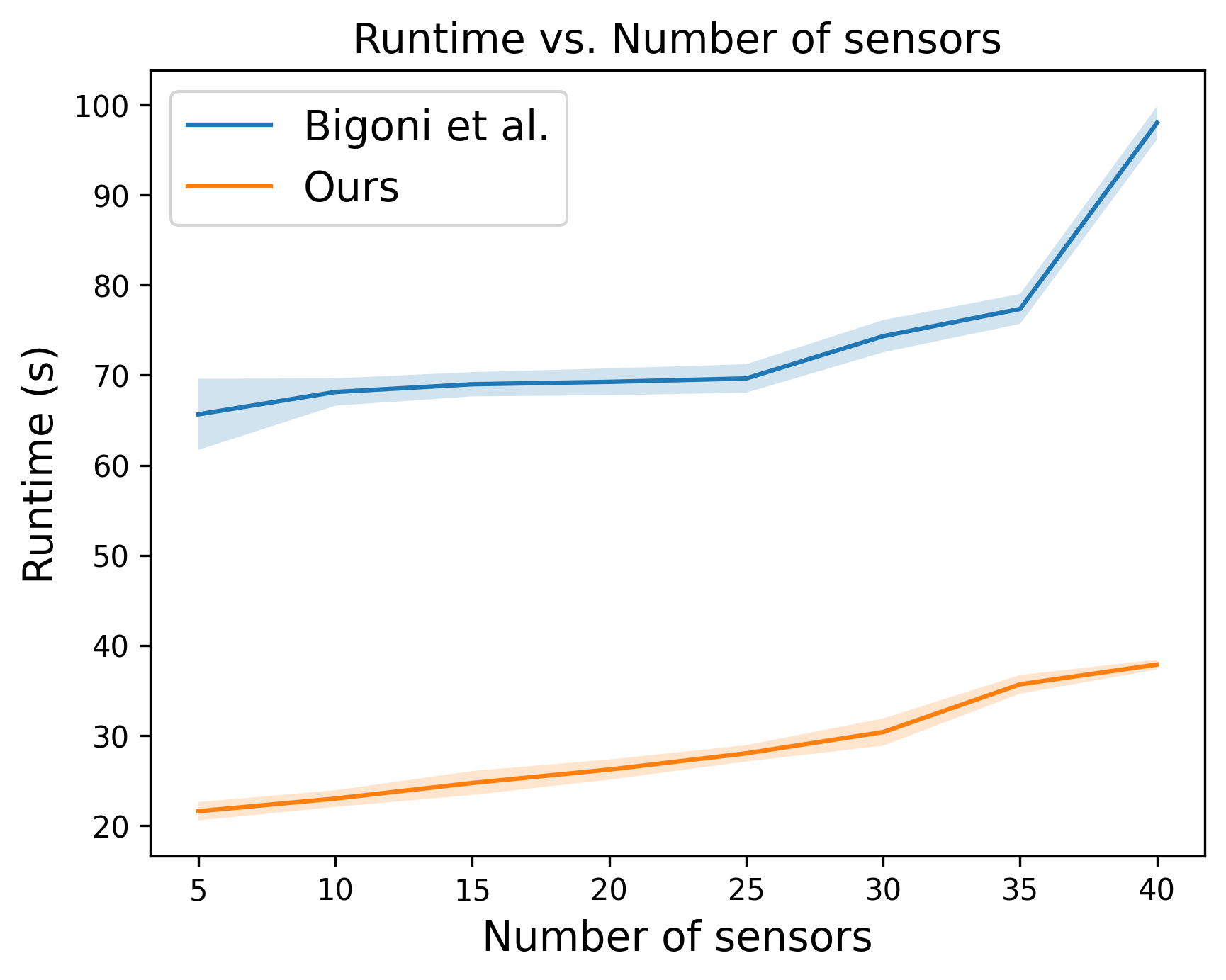}
        \subcaption{}
        \end{subfigure}
    \hspace{2cm}
    \begin{subfigure}{0.4\textwidth}
        \includegraphics[width=\textwidth]{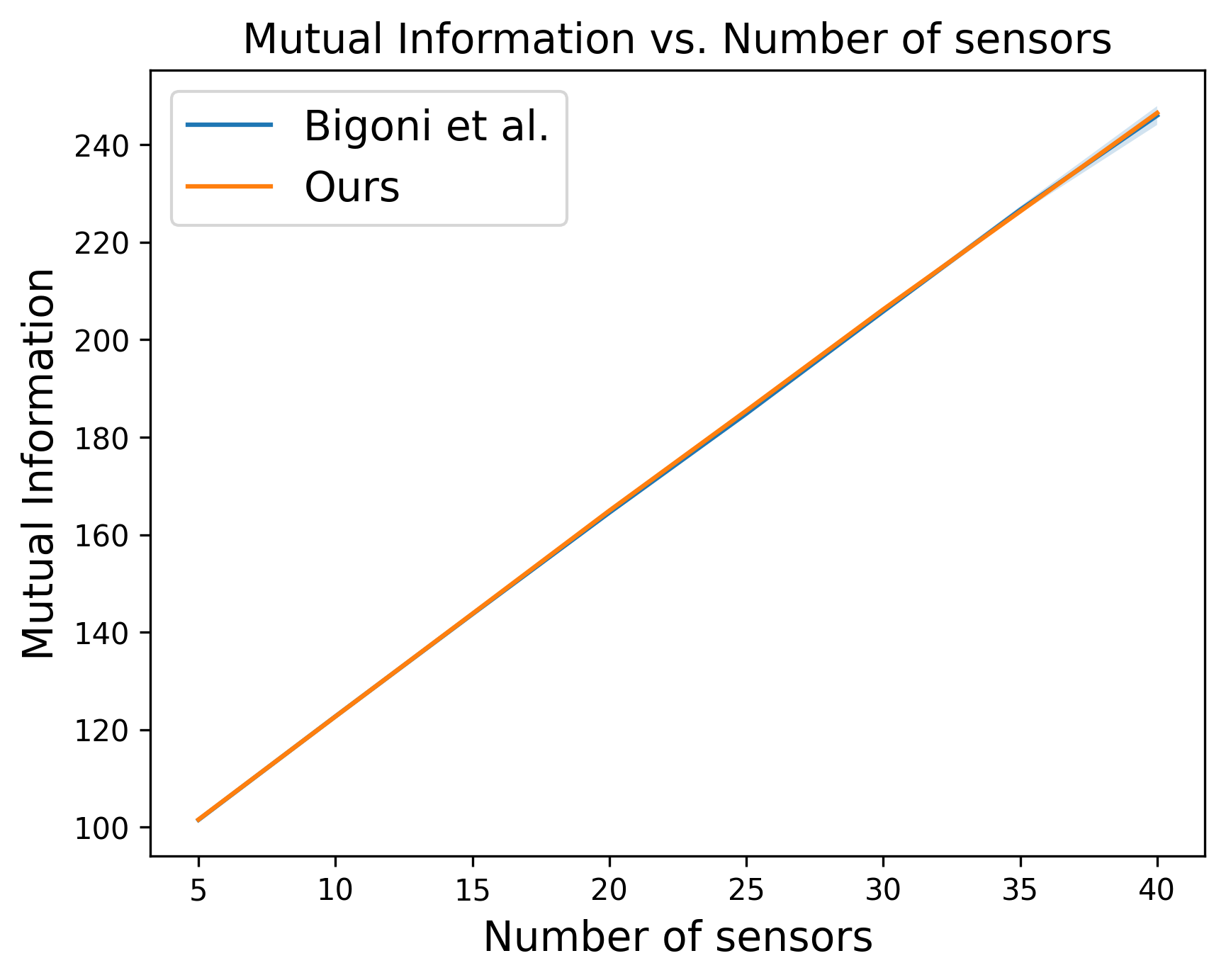}
        \subcaption{}
    \end{subfigure}
    \caption{Mean and standard deviation of runtime and MI of the solution placements from \citet{BigoniZH20}, and our Continuous-SGP approach (averaged over 10 runs of the experiment).}
    \label{fig:Bigonietal-benchmark}
\end{figure}

\begin{figure}[!hb]
    \centering
   \centering
    \begin{subfigure}{0.9\textwidth}
        \includegraphics[width=\textwidth]{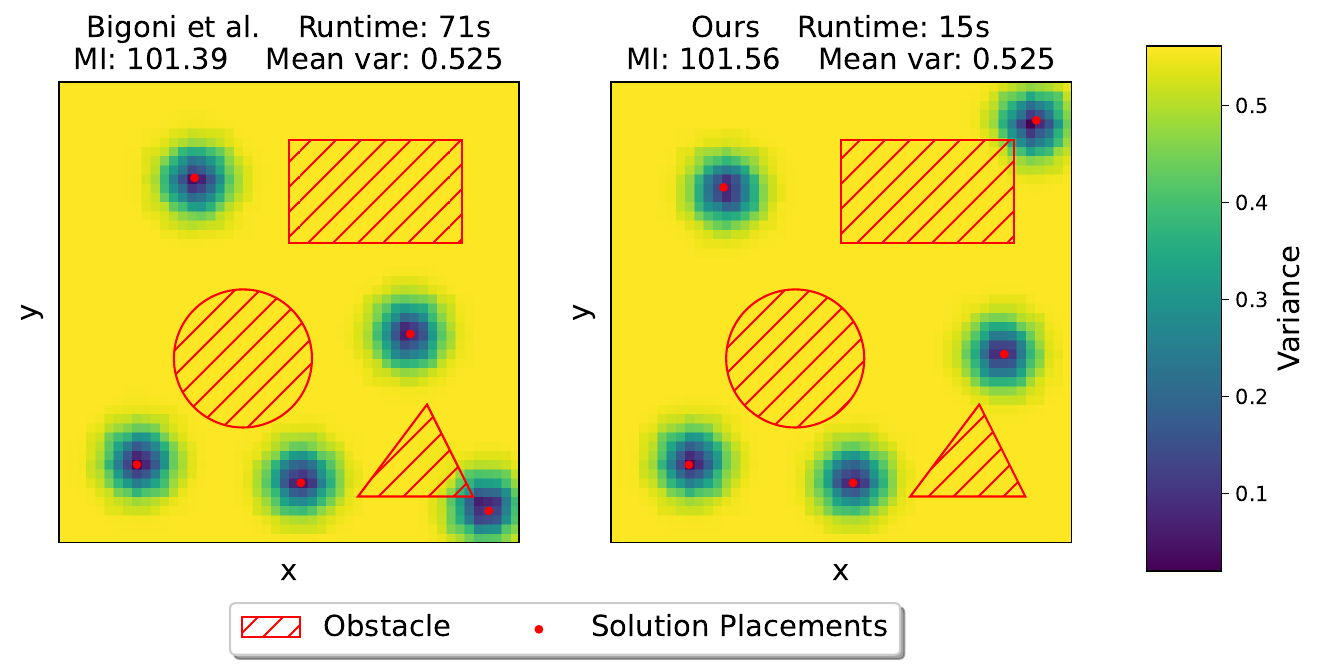}
        \subcaption{5 sensor placements}
        \end{subfigure}
    \hspace{2cm}
    \begin{subfigure}{0.9\textwidth}
        \includegraphics[width=\textwidth]{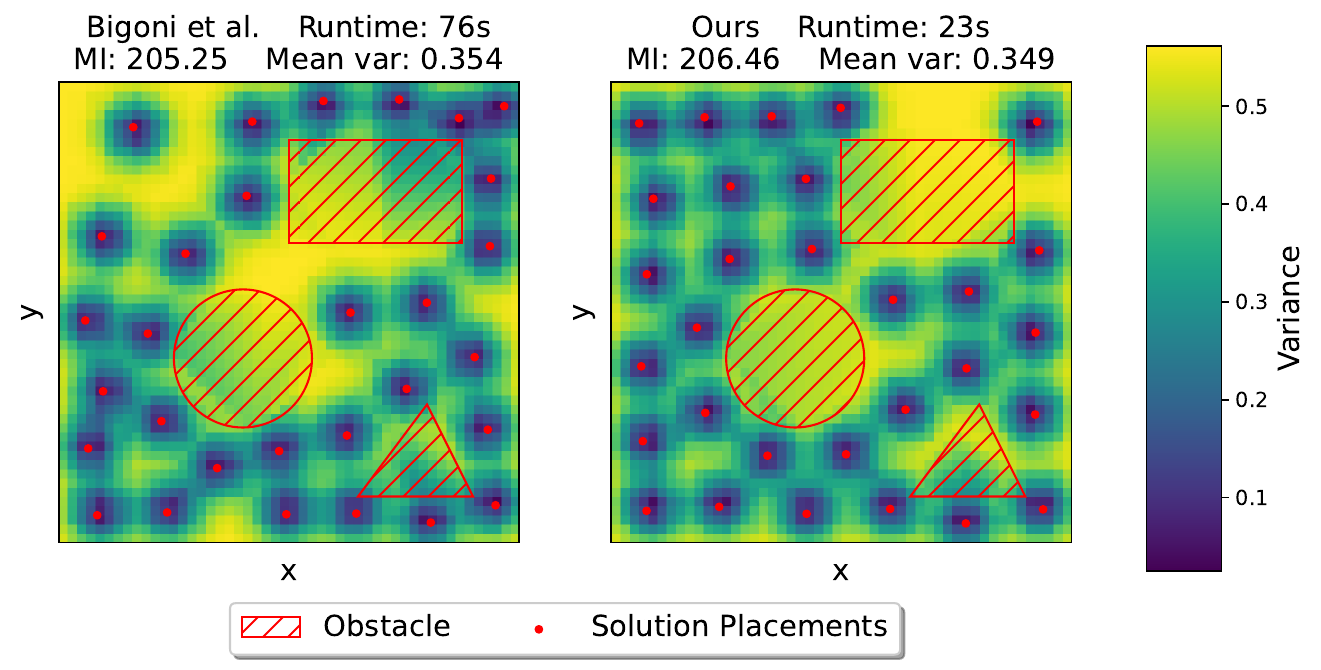}
        \subcaption{30 sensor placements}
    \end{subfigure}
    \caption{Solution sensor placements and estimated variance (uncertainty) at each location in the environment. (left) Solution placements from \citet{BigoniZH20}, (right) solution placements from our Continuous-SGP approach. Note that for mutual information (MI), higher is better, and for mean prediction variance (Mean var), lower is also better. Note that our SGP approach is considerably faster than the baseline.}
    \label{fig:Bigonietal}
\end{figure}

We benchmarked the two methods on an environment modeled with an RBF kernel with a length scale of 12.33 and variance of 0.56. We optimized the solutions for 5 to 40 sensor placements in increments of 5. Figure~\ref{fig:Bigonietal-benchmark} plots the optimization runtime and the mutual information between the solutions from the two approaches against the number of sensor placements, and Figure~\ref{fig:Bigonietal} shows the solution placements from the two approaches for 30 sensor placements. As we can see, our gradient descent-based approach is significantly faster than the baseline.

\clearpage
\section{Main Experiment Details and Results}
\subsection{Experiment Setup}
We used an RBF kernel~\citep{RasmussenW05} in all our experiments included in the main paper, and trained all GPs (and SGPs) with a learning rate of $0.01$ for a maximum of $3000$ iterations using the Adam optimizer~\citep{KingmaB14}. We used the GPflow Python library~\citep{WilkDJAAH20} for all our GP implementations, and the apricot Python library~\citep{SchreiberBN20} for the greedy selection algorithm.

All our experiments were executed on a Dell workstation with an Intel(R) Xeon(R) W-2265 CPU and 128 GB RAM. We ran our experiments using Python 3.8.10.

\subsection{KL Divergence Results}
\begin{figure}[h]
   \centering
    \begin{subfigure}[b]{0.24\textwidth}
        \includegraphics[width=\textwidth]{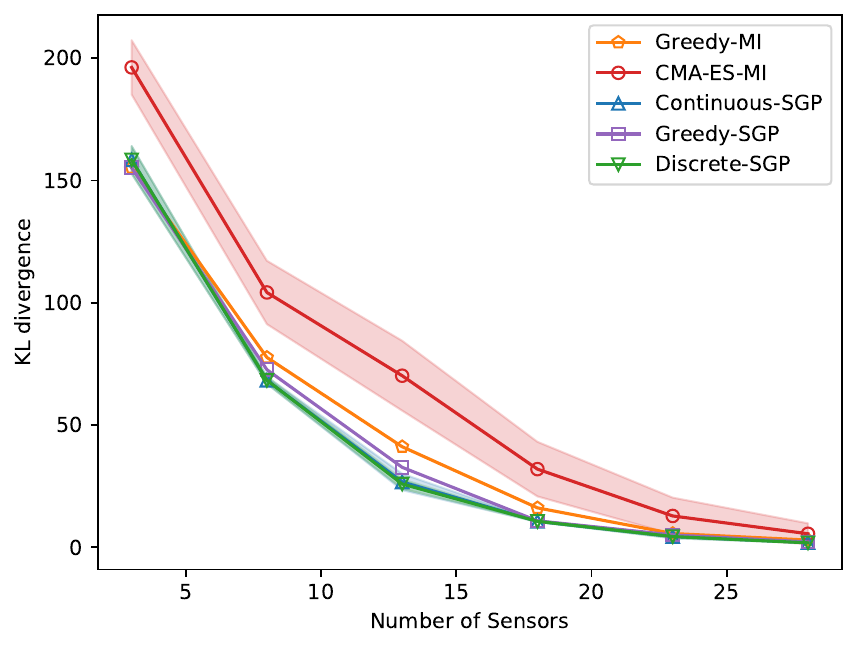}
        \caption{Temperature dataset}
        \label{fig:kl-intel}
    \end{subfigure}
    \hfill
    \begin{subfigure}[b]{0.24\textwidth}
        \includegraphics[width=\textwidth]{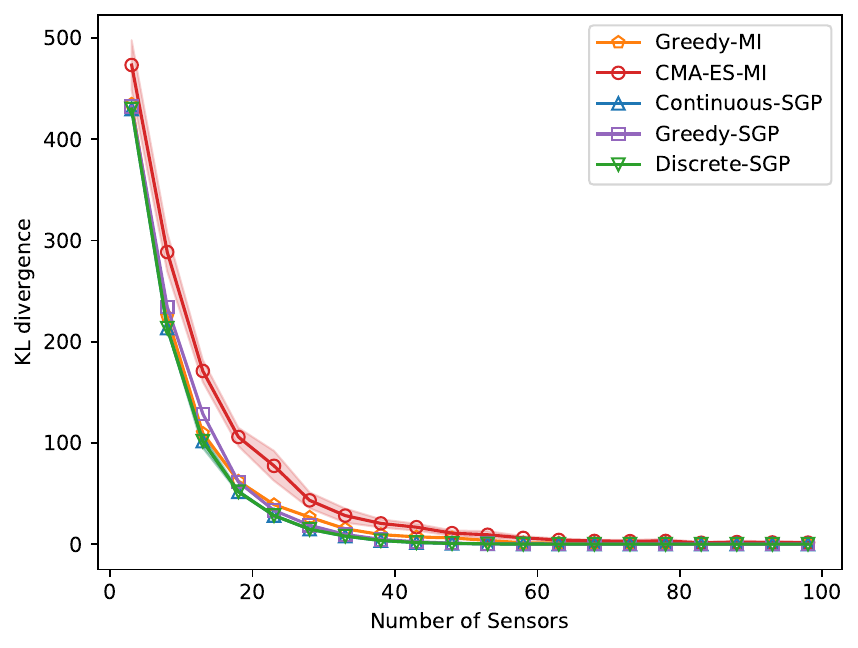}
        \caption{Precipitation dataset}
        \label{fig:kl-precip}
    \end{subfigure}
    \hfill
    \begin{subfigure}[b]{0.24\textwidth}
        \includegraphics[width=\textwidth]{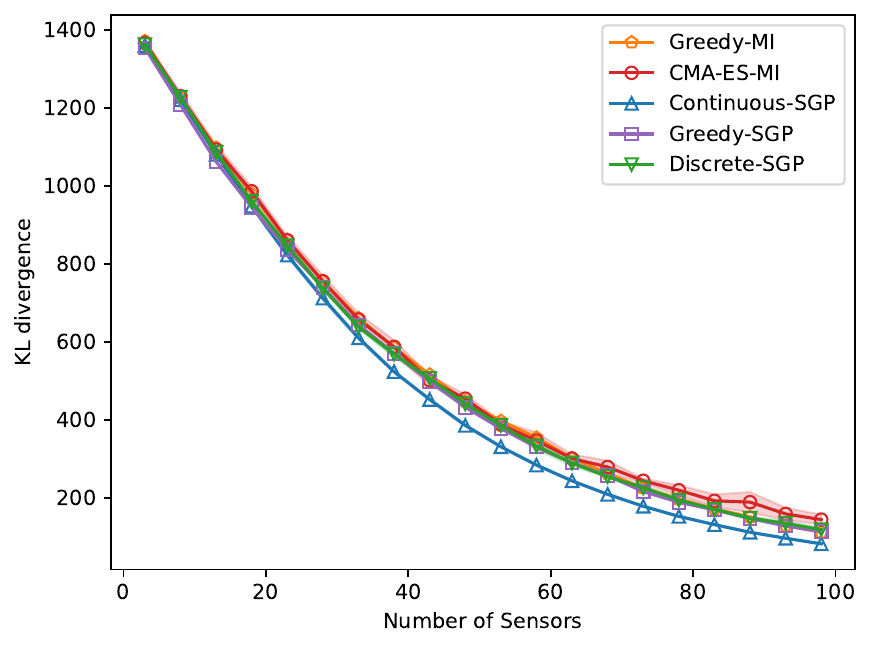}
        \caption{Soil dataset}
        \label{fig:kl-soil}
    \end{subfigure}
    \hfill
    \begin{subfigure}[b]{0.24\textwidth}
        \includegraphics[width=\textwidth]{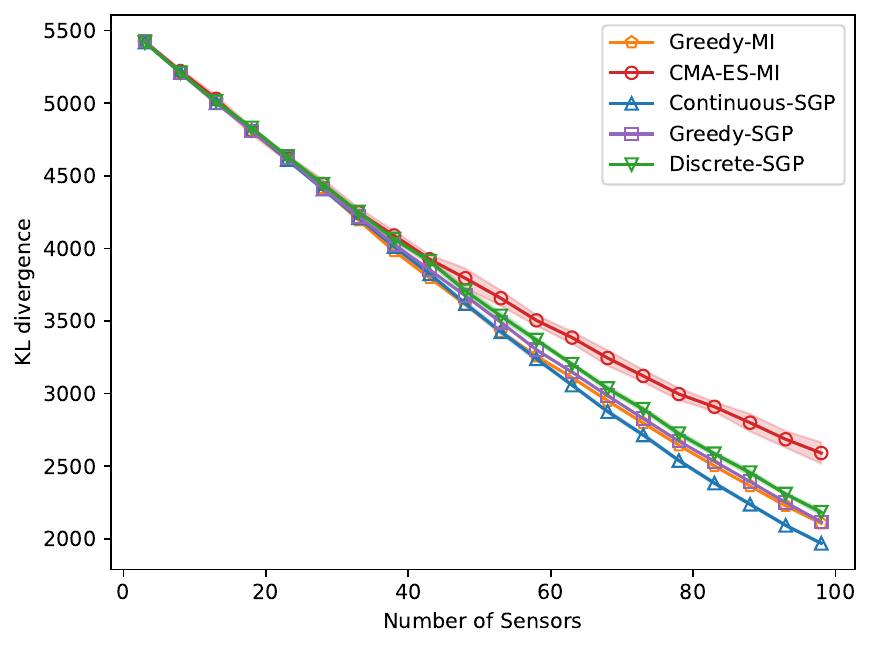}
        \caption{Salinity dataset}
        \label{fig:kl-sal}
    \end{subfigure}
    \caption{KL divergence between the SGP posterior and the true posterior vs number of sensors for the temperature, precipitation, soil, and salinity datasets (lower is better).}
    \label{fig:kl-benchmark}
\end{figure}

In Figure~\ref{fig:kl-benchmark}, we show how each of the methods in the paper—Greedy-MI~\citep{krauseSG08}, CMA-ES-MI~\citep{HitzGGPS17}, Continuous-SGP, Greedy-SGP, and Discrete-SGP—perform on the SGP's KL divergence measured using the approach presented in~\cite{BurtRW19}. We see the our approaches consistently perform on par or better than the baselines. 

\subsection{Bayesian Optimization Results}

\begin{figure}[h]
   \centering
    \begin{subfigure}[b]{0.24\textwidth}
        \includegraphics[width=\textwidth]{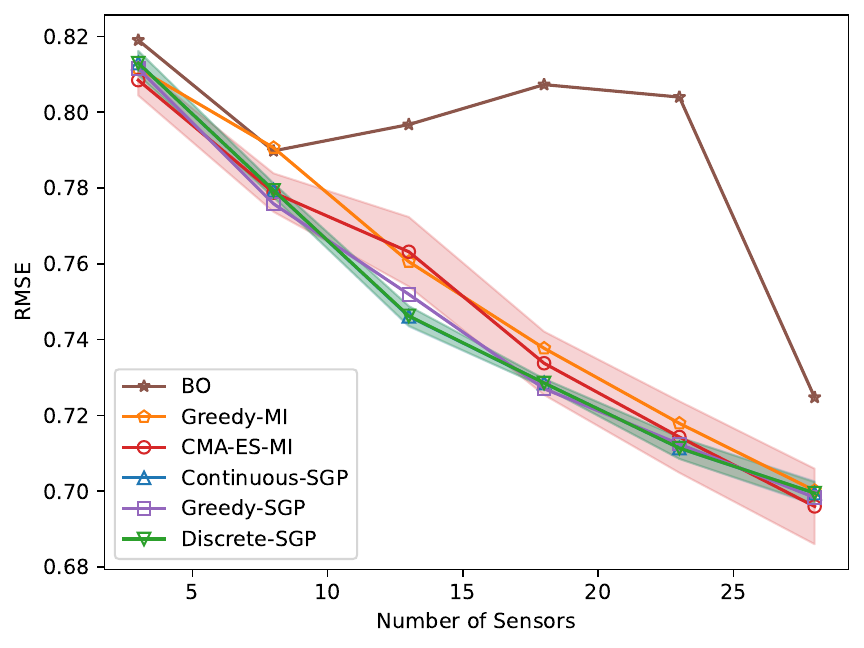}
        \caption{RMSE}
        \label{fig:rmse}
    \end{subfigure}
    \hfill
    \begin{subfigure}[b]{0.24\textwidth}
        \includegraphics[width=\textwidth]{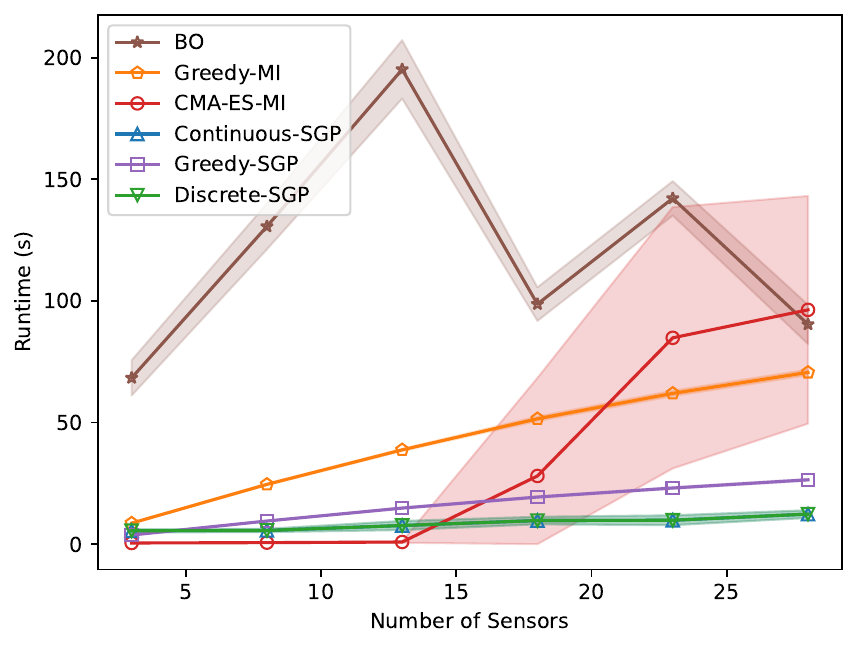}
        \caption{Runtime}
        \label{fig:runtime}
    \end{subfigure}
    \hfill
    \begin{subfigure}[b]{0.24\textwidth}
        \includegraphics[width=\textwidth]{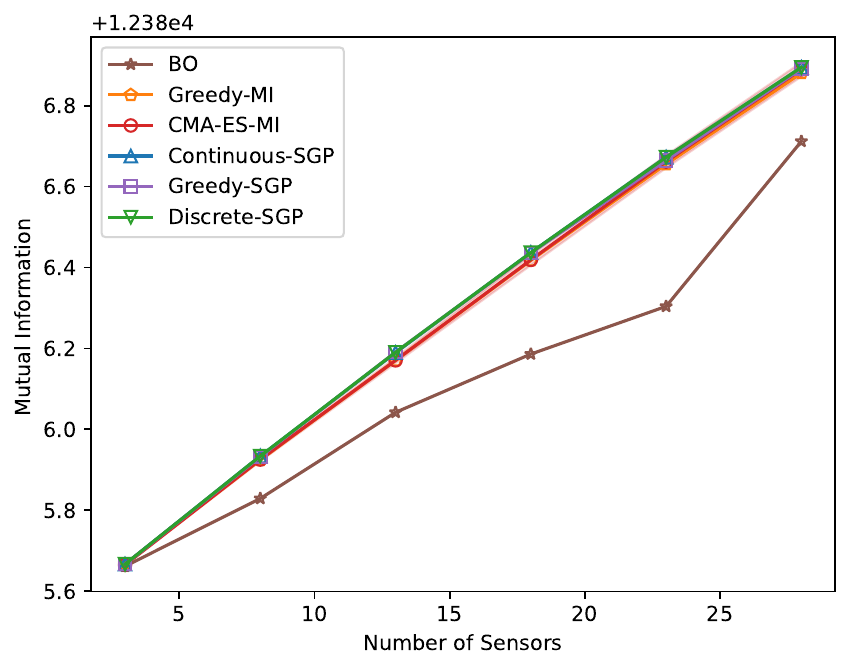}
        \caption{Mutual Information}
        \label{fig:mi}
    \end{subfigure}
    \hfill
    \begin{subfigure}[b]{0.24\textwidth}
        \includegraphics[width=\textwidth]{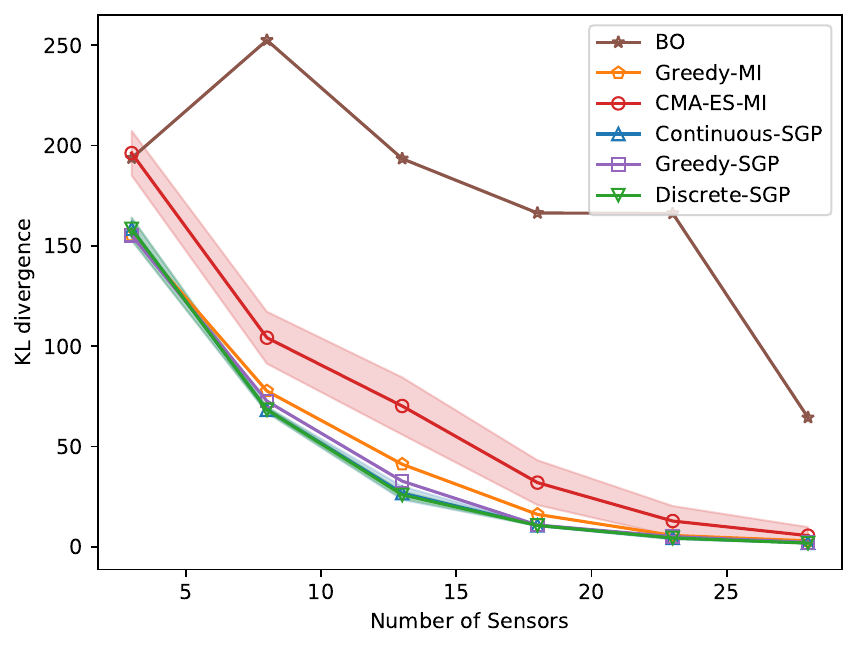}
        \caption{KL Divergence}
        \label{fig:kl}
    \end{subfigure}
    \caption{Results on the Intel temperature dataset generated using approach in~\citet{FrancisOMR19} and the methods in the paper—Greedy-MI~\citep{krauseSG08}, CMA-ES-MI~\citep{HitzGGPS17}, Continuous-SGP, Greedy-SGP, and Discrete-SGP.}
    \label{fig:bo-benchmark}
\end{figure}

For completeness, we also generated results on the Intel temperature dataset using the method in \citet{FrancisOMR19}, which maximizes GP-based MI using Bayesian optimization (BO), shown in Figure~\ref{fig:bo-benchmark}. However, the method's performance was subpar compared to the other baselines.

\subsection{Non-Stationarity Kernels}

Based on the experiments in Section~\ref{add_exp}, when using stationary kernels, the solution sensor placements appear to be uniformly distributed. Indeed, the benefit of the method is in determining the density of the sensing locations and avoiding the obstacles. This is true for approaches that maximize MI~\citep{krauseSG08, HitzGGPS17, FrancisOMR19} and our SGP approach, which is also an efficient method to approximately maximize MI. One can also use a faster uniform sampling technique such as Latin hypercube sampling~\citep{MckayBC79} to achieve similar results if the ideal sensing density is known and there are no obstacles in the environment. 

\begin{figure*}[!hb]
   \centering
    \begin{subfigure}{0.24\textwidth}
        \includegraphics[width=\textwidth]{figures/results/non_stationary_demo_gt.pdf}
        \subcaption{}
        \end{subfigure}
    \hfill
    \begin{subfigure}{0.24\textwidth}
        \includegraphics[width=\textwidth]{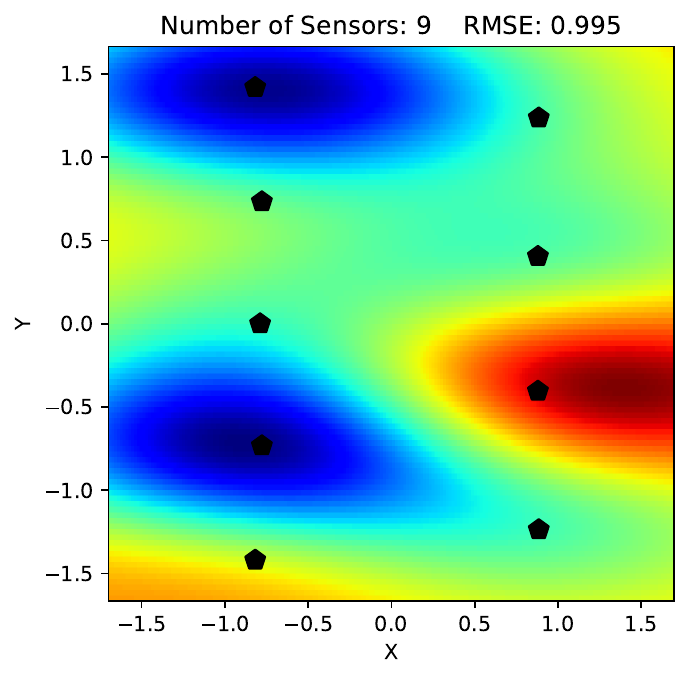}
        \subcaption{}
    \end{subfigure}
    \hfill
    \begin{subfigure}{0.24\textwidth}
        \includegraphics[width=\textwidth]{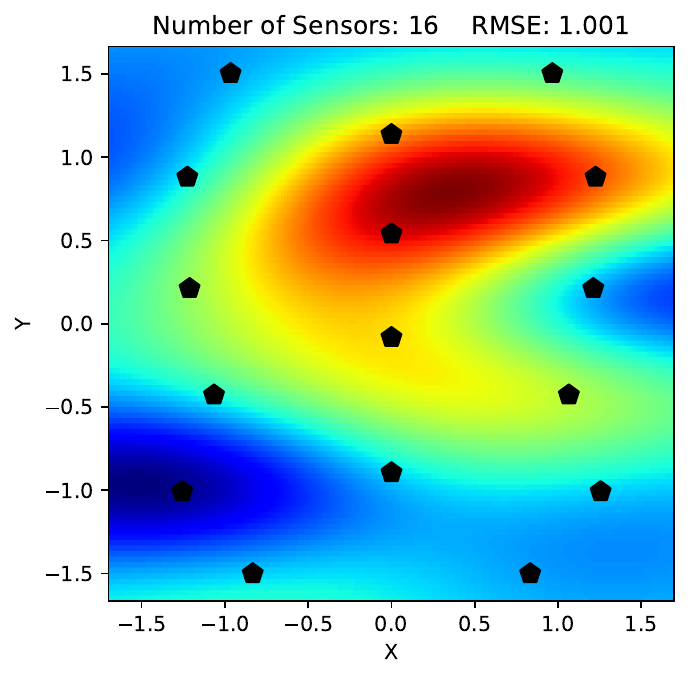}
        \subcaption{}
    \end{subfigure}
    \hfill
    \begin{subfigure}{0.24\textwidth}
        \includegraphics[width=\textwidth]{figures/results/stationary_demo_32.pdf}
        \subcaption{}
        \end{subfigure}
    \caption{A non-stationary environment. (a) Ground truth. Reconstructions from the Continuous-SGP solutions with a stationary RBF kernel function for (b) 9, (c) 16,  and (d) 32 sensing locations. The black pentagons represent the solution placements.}
    \label{fig:stat-benchmark}
\end{figure*}

However, when considering non-stationary environments, the key advantage of GP-based methods (including our SGP approach) becomes apparent. We generated a non-stationary data field (elevation data), in which the left half varies at a lower frequency than the right half. Figure~\ref{fig:stat-benchmark} shows the sensor placement results generated using the SGP approach with a stationary RBF kernel~\citep{RasmussenW05}. The ground truth data from the sensor placements was used to generate a dense reconstruction of the non-stationary environment (20,000 data points). As we can see, the method performs poorly at reconstructing the ground truth data. 

\begin{figure*}[!hb]
   \centering
    \begin{subfigure}{0.24\textwidth}
        \includegraphics[width=\textwidth]{figures/results/non_stationary_demo_gt.pdf}
        \subcaption{}
        \end{subfigure}
    \hfill
    \begin{subfigure}{0.24\textwidth}
        \includegraphics[width=\textwidth]{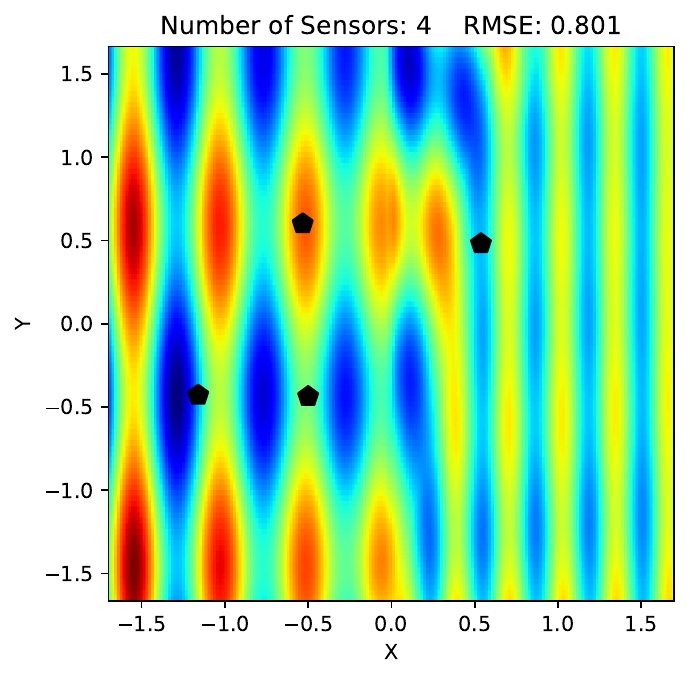}
        \subcaption{}
    \end{subfigure}
    \hfill
    \begin{subfigure}{0.24\textwidth}
        \includegraphics[width=\textwidth]{figures/results/non_stationary_demo_9.pdf}
        \subcaption{}
    \end{subfigure}
    \hfill
    \begin{subfigure}{0.24\textwidth}
        \includegraphics[width=\textwidth]{figures/results/non_stationary_demo_16.pdf}
        \subcaption{}
    \end{subfigure}
    \caption{A non-stationary environment. (a) Ground truth. Reconstructions from the Continuous-SGP solutions with a neural kernel for (b) 4, (c) 9 and, (d) 16 sensing locations. The black pentagons represent the solution placements.}
    \label{fig:non-stat-benchmark-supp}
\end{figure*}

We can address the above issue by leveraging a non-stationary kernel. We trained a neural kernel~\citep{RemesHK18} to learn the correlations in the environment. We used three mixture components in the neural kernel function and parameterized each constituent neural network as a two-layer multilayer perceptron~\citep{Bishop06} with four hidden units each. The training data consisted of 1250 grid-sampled labeled data from the non-stationary environment. The neural kernel function parameters were optimized using a Gaussian process (GP)~\citep{RasmussenW05} trained with type-II maximum likelihood~\citep{Bishop06}. We then used the neural kernel function in our Continuous-SGP approach~(Algorithm~\ref{alg:Continuous-SGP}) to generate sensor placements for 4, 9, and 16 sensing locations. The results with the non-stationary neural kernel are shown in Figure~\ref{fig:non-stat-benchmark-supp} for 4, 9, and 16 sensor placements. 

As we can see in Figure~\ref{fig:non-stat-benchmark-supp}, the sensor placements are no longer uniformly distributed. Instead, they are biased towards strategic locations that are crucial for accurately estimating the data field. We see multiple sensors in the middle, where the frequency of the variations (elevation) change, thereby accurately capturing the underlying data field.

\end{document}